\documentclass[reviewcopy]{elsart}

\usepackage{tabularx}
\usepackage{subfigure}

\usepackage{times}
\usepackage{graphicx}
\usepackage{array}
\usepackage{amsmath}
\usepackage{amssymb}
\usepackage{algorithm}
\usepackage{algorithmic}
\usepackage{url}
\usepackage{multirow}

\newcolumntype{V}{>{$\vcenter\bgroup\hbox\bgroup}c<{\egroup\egroup$}}

\graphicspath{{imgs/}}

\begin{document}

\begin{frontmatter}

\title{Hallucinating Optimal High-Dimensional Subspaces}

\author[DEAK]{Ognjen Arandjelovi\'c}

\address[DEAK]{Centre for Pattern Recognition and Data Analytics (PRaDA)\\School of Information Technology\\Deakin University\\Geelong 3216 VIC\\Australia\\~\\
               Tel: +61(0)3522-73079\\E-mail: \texttt{ognjen.arandjelovic@gmail.com}\\Web: \texttt{http://mi.eng.cam.ac.uk/$\sim$oa214} }

\title{}

\author{}

\address{}

\clearpage

~\\
\begin{abstract}
Linear subspace representations of appearance variation are pervasive in
computer vision. This paper addresses the problem of robustly matching
such subspaces (computing the similarity between them) when they are used
to describe the scope of variations within sets of images of different
(possibly greatly so) scales. A na\"{\i}ve solution of projecting the
low-scale subspace into the high-scale image space is described first
and subsequently shown to be inadequate, especially at large scale discrepancies.
A successful approach is proposed instead. It consists of (i) an interpolated
projection of the low-scale subspace into the high-scale space, which is
followed by (ii) a rotation of this initial estimate within the bounds of
the imposed ``downsampling constraint''. The optimal rotation is found in
the closed-form which best aligns the high-scale reconstruction of the
low-scale subspace with the reference it is compared to. The method is
evaluated on the problem of matching sets of (i) face appearances under
varying illumination and (ii) object appearances under varying viewpoint,
using two large data sets. In comparison to the na\"{\i}ve matching, the
proposed algorithm is shown to greatly increase the separation of
between-class and within-class similarities, as well as produce far more
meaningful modes of common appearance on which the match score is based.
\end{abstract}


\end{frontmatter}

\clearpage

\section{Introduction}\label{s:intro}
One of the most commonly encountered problems in computer vision is
that of matching appearance. Whether it is images of local features
\cite{FerrTuytVanG2004}, views of objects
\cite{EverZissWillVanG2006} or faces \cite{SuShanChenGao2009},
textures \cite{PradBhutNasiPrad2009} or rectified planar structures
(buildings, paintings) \cite{HartZiss2004}, the task of comparing
appearances is virtually unavoidable in a modern computer vision
application. A particularly interesting and increasingly important
instance of this task concerns the matching of \emph{sets} of
appearance images, each set containing examples of variation
corresponding to a single class.

A ubiquitous representation of appearance variation within a class
is by a linear subspace \cite{ChenSute2006,Beth2006}. The most basic
argument for the linear subspace representation can be made by
observing that in practice the appearance of interest is constrained
to a small part of the image space. Domain-specific information may
restrict this even further e.g.\ for Lambertian surfaces seen from a
fixed viewpoint but under variable illumination
\cite{BelhKrie1998,GeorBelhKrie2001,BasrJaco2003} or smooth objects
across changing pose \cite{LeeHoYangKrie2005,ZhouAggarChell+2007}.
What is more, linear subspace models are also attractive for their
low storage demands -- they are inherently compact and can be learnt
incrementally
\cite{SkocLeon2008,SongWang2005,AranCipo2005a,VerbVlasKros2003,HallMarsMart2000,GrosYangWaib2000}.
Indeed, throughout this paper it is assumed that the original data
from which subspaces are estimated is not available.

A problem which arises when trying to match two subspaces -- each
representing certain appearance variation -- and which has not as of
yet received due consideration in the literature, is that of
matching subspaces embedded in different image spaces, that is,
corresponding to image sets of different scales. This is a frequent
occurrence: an object one wishes to recognize may appear larger or
smaller in an image depending on its distance, just as a face may,
depending on the person's height and positioning relative to the
camera. In most matching problems in the computer vision literature,
this issue is overlooked. Here it is addressed in detail and shown that
a na\"{\i}ve approach to normalizing for scale in subspaces results
in inadequate matching performance. Thus, a method is proposed which
\emph{without any assumptions on the nature of appearance} that the
subspaces represent, constructs an optimal hypothesis for a
high-resolution reconstruction of the subspace corresponding to
low-resolution data.

In the next section, a brief overview of the linear subspace
representation is given first, followed by a description of the
aforementioned na\"{\i}ve scale normalization. The proposed solution
is described in this section as well. In Section~\ref{s:exp} the
two approaches are compared empirically and the results analyzed
in detail. The main contribution and conclusions of the paper are
summarized in Section~\ref{s:conclusion}.

\section{Matching Subspaces across Scale}\label{s:matching}
Consider a set $X \subset \mathbb{R}^{d}$ containing vectors which
represent rasterized images:
\begin{align}
  &X = \big\{ \mathbf{x}_1, \ldots, \mathbf{x}_N \big\}
\end{align}
where $d$ is the number of pixels in each image. It is assumed that
all of the images represented by members of $X$ have the same aspect
ratio, so that the same indices of different vectors correspond
spatially to the same pixel location. A common representation of
appearance variation described by $X$ is by a linear subspace of
dimension $D$, where usually it is the case that $D \ll d$. If
$\mathbf{m}_X$ is the estimate of the mean of the samples in $X$:
\begin{align}
  \mathbf{m}_X &= \frac{1}{N}   \sum_{i=1}^N \mathbf{x}_i,
\end{align}
then $\mathbf{B}_X \in \mathbb{R}^{d \times D}$, a matrix with
columns consisting of orthonormal basis vectors spanning the
$D$-dimensional linear subspace embedded in a $d$-dimensional image
space, can be computed from the corresponding covariance matrix
$\mathbf{C}_X$:
\begin{align}
  \mathbf{C}_X &= \frac{1}{N-1} \sum_{i=1}^N \big(\mathbf{x}_i - \mathbf{m}_X\big) \big(\mathbf{x}_i - \mathbf{m}_X\big)^T.
\end{align}
Specifically, an insightful interpretation of $\mathbf{B}_X$ is as
the row and column space basis of the best rank-$D$ approximation to
$\mathbf{C}_X$:
\begin{align}
  \mathbf{B}_X = \arg \min_{\scriptsize
  \begin{array}{c}
    \mathbf{B} \in \mathbb{R}^{d \times D}\\
    \mathbf{B}^T\mathbf{B} = I
    \end{array}} \min_{\scriptsize
  \begin{array}{c}
    \Lambda \in \mathbb{R}^{D\times D}\\
    \Lambda_{ij} = 0, i \neq j
    \end{array}}
  {\big\|~\mathbf{C}_X - \mathbf{B}~\Lambda~\mathbf{B}^T~\big\|_F}^2,
\end{align}
where $\| . \|_F$ is the Frobenius norm of a matrix.

\subsection{The ``Na\"{\i}ve Solution''}\label{ss:naive}
Let $\mathbf{B}_X \in \mathbb{R}^{d_l \times D}$ and $\mathbf{B}_Y
\in \mathbb{R}^{d_h \times D}$ be two basis vectors matrices
corresponding to appearance variations of image sets containing
images with $d_l$ and $d_h$ pixels respectively. Without loss of
generality, let also $d_l < d_h$. As before, here it is assumed that
all images both within each set, as well as across the two sets, are
of the same aspect ratio. Thus, we wish to compute the similarity of sets
represented by orthonormal basis matrices $\mathbf{B}_X$ and
$\mathbf{B}_Y$.

\begin{figure}[htb]
  \centering
  \vspace{20pt}
  \subfigure[]{\includegraphics[width=1.00\textwidth]{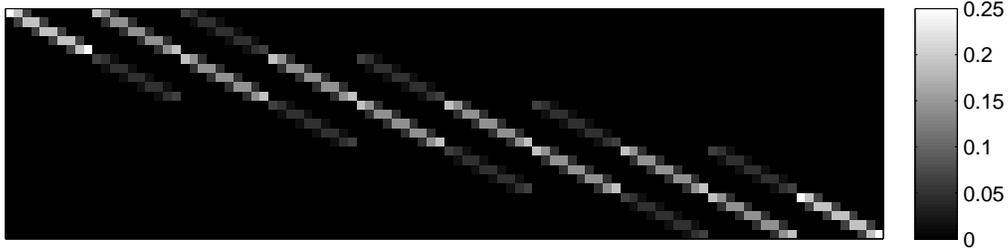}}
  \subfigure[]{\includegraphics[width=1.00\textwidth]{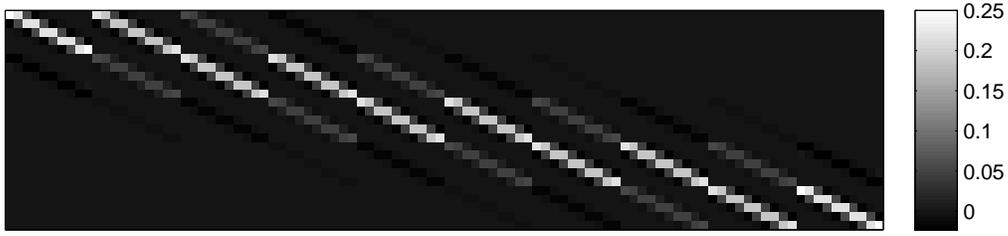}}
  \caption{ The projection matrix $\mathbf{P} \in \mathbb{R}^{25 \times 100}$ modelling the process of downsampling
            a $10 \times 10$ pixel image to $5 \times 5$ pixels, using (a) bilinear and (b) bicubic projection models,
            shown as an image. For the interpretation of image intensities see the associated grey level scales
            on the right. }
  \label{f:pmat}
  \vspace{20pt}
\end{figure}

Subspaces spanned by the columns of $\mathbf{B}_X$ and
$\mathbf{B}_Y$ cannot be compared directly as they are embedded in
different image spaces. Instead, let us model the process of an
isotropic downsampling of a $d_h$-pixel image down to $d_l$ pixels
with a linear projection realized though a projection matrix
$\mathbf{P} \in \mathbb{R}^{d_l \times d_h}$. In other words, for a
low-resolution image set $X \subset \mathbb{R}^{d_l}$:
\begin{align}
  &X = \big\{ \mathbf{x}_1, \ldots, \mathbf{x}_N \big\}
\end{align}
there is a high-resolution set $X^* \subset \mathbb{R}^{d_h}$, such
that:
\begin{align}
  &X^* = \bigg\{ \mathbf{x}_i^*~|~\mathbf{x}_i = \mathbf{P}~\mathbf{x}_i^*~;~i=1,\ldots,N \bigg\}.
\end{align}
The form of the projection matrix depends on (i) the projection model employed (e.g.\ bilinear, bicubic etc.)
and (ii) the dimensions of high and low scale images; see Figure~\ref{f:pmat} for an illustration.

Under the assumption of a linear projection model, the least-square error reconstruction of the
high-dimensional data can be achieved with a linear projection as well, in this case by $\mathbf{P}_R$
which can be computed as:
\begin{align}
  \mathbf{P}_R = \mathbf{P}^T~\left(\mathbf{P}~\mathbf{P}^T\right)^{-1}.
\end{align}
Since it is assumed that the original data from which $\mathbf{B}_X$ was
estimated is not available, an estimate of the subspace
corresponding to $X^*$ can be computed by re-projecting each of the
basis vectors (columns) of $\mathbf{B}_X$ into $\mathbb{R}^{d_h}$:
\begin{align}
  \tilde{\mathbf{B}}^*_X = \mathbf{P}_R~\mathbf{B}_X.
\end{align}
Note that in general $\tilde{\mathbf{B}}^*_X$ is not an orthonormal matrix \textit{i.e}.\
${{}\tilde{\mathbf{B}}_X^*}^T~\tilde{\mathbf{B}}^*_X \neq \mathbf{I}$. Thus, after re-projecting the
subspace basis, it is orthogonalized using the Householder transformation \cite{Hous1958},
producing a high-dimensional subspace basis estimate $\mathbf{B}^*_X$ which can be compared directly with
$\mathbf{B}_Y$.

\subsubsection{Limitations of the Na\"{\i}ve Solution} The process of downsampling an image inherently
causes a loss of information. In re-projecting the subspace basis vectors, information gaps
are ``filled in'' through interpolation. This has the effect of constraining the spectrum of
variation in the high-dimensional reconstructions to the bandwidth of the low-dimensional data.
Compared to the genuine high-resolution images, the reconstructions are void of high frequency detail
which usually plays a crucial role in discriminative problems.

\subsection{Proposed Solution}\label{ss:proposed}
\begin{figure}[htb]
  \centering
  \includegraphics[width=0.95\textwidth]{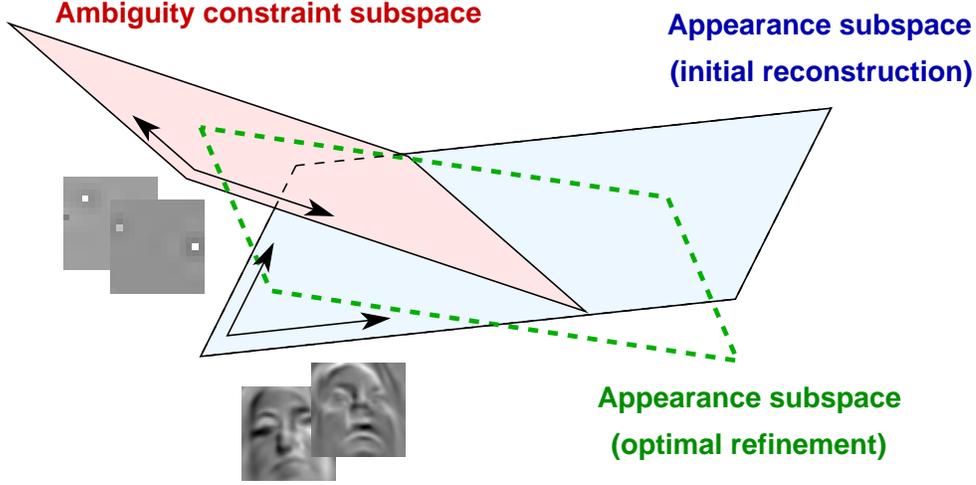}
  \caption{ A conceptual illustration of the main idea: the initial reconstruction of the class subspace in the
            high dimensional image space is refined through rotation within the constraints of the
            ambiguity constraint subspace. }
  \label{f:main}
  \vspace{20pt}
\end{figure}

We seek a constrained correction to the subspace basis
$\mathbf{B}^*_X$. To this end, consider a vector $\mathbf{x}^*_i$ in
the high-dimensional image space $\mathbb{R}^{d_h}$, which when
downsampled maps onto $\mathbf{x}_i$ in $\mathbb{R}^{d_l}$. As
before, this is modelled as a linear projection effected by a
projection matrix $\mathbf{P}$:
\begin{align}
  \mathbf{x}_i = \mathbf{P}~\mathbf{x}^*_i.
\end{align}
Writing the reconstruction of $\mathbf{x}^*_i$, computed as described in the previous section,
as $\mathbf{x}^*_i + \mathbf{c}_i$, it has to hold:
\begin{align}
  \mathbf{x}_i = \mathbf{P}~\big(\mathbf{x}^*_i + \mathbf{c}_i \big),
\end{align}
or, equivalently:
\begin{align}
  \mathbf{0} = \mathbf{P}~\mathbf{c}_i,
\end{align}
In other words, the correction term $\mathbf{c}_i$ has to lie in the
nullspace of $\mathbf{P}$. Let $\mathbf{B}_c$ be a matrix of basis
vectors spanning the nullspace which, given its meaning in the
proposed framework, will henceforth be referred to as the \emph{ambiguity
constraint subspace}. Then the actual appearance in the
high-dimensional image space corresponding to the subspace
$\mathbf{B}_X \in \mathbb{R}^{d_l \times D}$ is not spanned by the
$D$ columns of $\mathbf{B}^*_X$ but rather some $D$ orthogonal
directions in the span of the columns of
$\big[\mathbf{B}^*_X~|~\mathbf{B}_c \big]$, as illustrated in
Figure~\ref{f:main}.

Let $\mathbf{B}_{Xc}$ be a matrix of orthonormal basis vectors computed by orthogonalizing $\big[\mathbf{B}^*_X~|~\mathbf{B}_c \big]$:
\begin{align}
  \mathbf{B}_{Xc} = \text{orth} \big( \big[\mathbf{B}^*_X~|~\mathbf{B}_c \big] \big)
\end{align}
Then we seek a matrix $\mathbf{T} \in \mathbb{R}^{(D+d_h-d_l) \times
D}$ which makes the optimal choice of $D$ directions from the span
of $\mathbf{B}_{Xc}$:
\begin{align}
  \mathbf{B}_{Xc}~\mathbf{T} = \mathbf{B}_{Xc}~ \bigg[~\mathbf{t}_1~|~\mathbf{t}_2~|~\ldots~|~\mathbf{t}_D~\bigg].
\end{align}
Here the optimal choice of $\mathbf{T}$ is defined as the one that
best aligns the reconstructed subspace with the subspace it is
compared with, \textit{i.e}.\ $\mathbf{B}_Y$. The matrix $\mathbf{T}$ can be
constructed recursively, so let us consider how its first column
$\mathbf{t}_1$ can be computed. The optimal alignment criterion can
be restated as:
\begin{align}
  \mathbf{t}_1 &= \text{arg} \max_{\mathbf{t}'_1} \max_{\mathbf{a}}
    \frac{\big(\mathbf{B}_Y~\mathbf{a}\big) \cdot \big(\mathbf{B}_{Xc}~\mathbf{t}'_1\big)} {\| \mathbf{a} \|~\| \mathbf{t}'_1 \|}
    \label{e:corr}
\end{align}
Rewriting the right-hand side:
\begin{align}
  \mathbf{t}_1 &= \text{arg} \max_{\mathbf{t}'_1} \max_{\mathbf{a}}
    \frac{\big(\mathbf{B}_Y~\mathbf{a}\big) \cdot \big(\mathbf{B}_{Xc}~\mathbf{t}'_1\big)} {\| \mathbf{a} \|~\| \mathbf{\mathbf{t}'_1} \|}=\\
  &=\text{arg} \max_{\mathbf{t}'_1} \max_{\mathbf{a}}
    \frac{\mathbf{a}^T} {\| \mathbf{a} \|}~{\mathbf{B}_Y}^T~\mathbf{B}_{Xc}~\frac{\mathbf{t}'_1}{\| \mathbf{\mathbf{t}'_1} \|} =\\
  &=\text{arg} \max_{\mathbf{t}'_1} \max_{\mathbf{a}}
    \frac{\mathbf{a}^T} {\| \mathbf{a} \|}~\mathbf{U}~\Sigma~{\mathbf{V}}^T~\frac{\mathbf{t}'_1}{\| \mathbf{\mathbf{t}'_1} \|},
  \label{e:opt}
\end{align}
where:
\begin{align}
  {\mathbf{B}_Y}^T~&\mathbf{B}_{Xc} = \notag \\
  &\overbrace{\bigg[~\mathbf{u}_1~|~\ldots~|~\mathbf{u}_D~\bigg]}^{\mathbf{U}}~
   \overbrace{\left[
     \begin{array}{cccccc}
       ~~\sigma_1~~ & 0        & \ldots & 0        & \ldots & 0\\
       0        & ~~\sigma_2~~ & \ldots & 0        & \ldots & 0\\
       \vdots   & \vdots       & ~~\ddots~~ & \vdots   & \ldots & 0\\
       0        & 0            & \ldots & ~~\sigma_D~~ & \ldots & 0\\
     \end{array}\right]}^{\Sigma}~\overbrace{\bigg[~\mathbf{v}_1~|~\ldots~|~\mathbf{v}_{D+d_h-d_l}~\bigg]^T}^{\mathbf{V}^T},
\end{align}
is the Singular Value Decomposition (SVD) of ${\mathbf{B}_Y}^T~\mathbf{B}_{Xc}$ and $\sigma_1 \geq \sigma_2 \geq \ldots \geq \sigma_D$. Then, from the right-hand side in
Equation~\eqref{e:opt}, by inspection the optimal directions of $\mathbf{a}$ and $\mathbf{t}'_1$ are, respectively, the first SVD ``output'' direction $\mathbf{u}_1$ and the first SVD ``input'' direction, \textit{i.e}.\ $\mathbf{a}=\mathbf{u}_1$, and $\mathbf{t}_1 = \mathbf{v}_1$. The same process can be used to infer the remaining columns of $\mathbf{T}$, the $i$-th one being $\mathbf{t}_i = \mathbf{v}_i$.


Thus, the optimal reconstruction $\mathbf{B}'_X$ of $\mathbf{B}_{X}$
in the high-dimensional space, obtained by the constrained rotation
of the na\"{\i}ve estimate $\mathbf{B}^*_X$, is given by the
orthonormal basis matrix:
\begin{align}
  \mathbf{B}'_X = \mathbf{B}_{Xc}~\bigg[~\mathbf{v}_1~|~\ldots~|~\mathbf{v}_D~\bigg]
\end{align}
The key steps of the algorithm are summarized in
Figure~\ref{f:summary}.

\begin{figure}[t]
  \centering
  \begin{list}{\labelitemi}{\leftmargin=0.1\textwidth}
    \vspace{25pt}
    \hrule
    \vspace{5pt}
    \item[] {\bf \hspace{-30pt}Input:}~~~~~Orthonormal subspace basis matrices $\mathbf{B}_X \in \mathbb{R}^{d_l}$, $\mathbf{B}_Y \in \mathbb{R}^{d_h}$\\
             \hspace{10pt}Projection model $\mathbf{P} \in \mathbb{R}^{d_l \times d_h}$\\
    \item[] {\bf \hspace{-30pt}Output:}~~Optimal reconstruction $\mathbf{B}'_X$ of the high-dimensional space\\
                \hspace{10pt}corresponding to $\mathbf{B}_X$\\\vspace{10pt}
    \hrule\vspace{10pt}
    \item[] {\hspace{-30pt}1: Compute the reverse projection matrix}~\newline
        $\mathbf{P}_R = \mathbf{P}^T~\left(\mathbf{P}~\mathbf{P}^T\right)^{-1}$

    \item[] {\hspace{-30pt}2: Compute the initial na\"{\i}ve reconstruction}~\newline
        $\mathbf{B}^*_X = \text{orth}\big(\mathbf{P}_R~\mathbf{B}_X\big)$

    \item[] {\hspace{-30pt}3: Compute a basis of the ambiguity constraint subspace}~\newline
        $\mathbf{B}_c = \text{nullspace}\big(\mathbf{P}\big)$

    \item[] {\hspace{-30pt}4: Compute a joint basis of the initial reconstruction and the ambiguity constraint subspace}~\newline
        $\mathbf{B}_{Xc} = \text{orth}\big(~[\mathbf{B}_c~|~\mathbf{B}^*_X]~\big)$

    \item[] {\hspace{-30pt}6: Perform Singular Value Decomposition of ${\mathbf{B}_Y}^T~\mathbf{B}_{Xc}$}~\newline
        ${\mathbf{B}_Y}^T~\mathbf{B}_{Xc} = \mathbf{U}~\Sigma~\mathbf{V}^T = \mathbf{U}~\Sigma~\big[ \mathbf{v}_1~|~\ldots~|~\mathbf{v}_D \big]^T$

    \item[] {\hspace{-30pt}7: Extract the orthonormal basis of the best reconstruction}~\newline
        $\mathbf{B}'_X = \mathbf{B}_{Xc}~\big[~\mathbf{v}_1~|~\ldots~|~\mathbf{v}_D~\big]$\vspace{10pt}
    \hrule
  \end{list}
  \vspace{10pt}
  \caption{A summary of the proposed matching algorithm.}
  \label{f:summary}
\end{figure}

\subsubsection{Computational Requirements and Implementation Issues}
Before turning our attention to the empirical analysis of the
proposed algorithm let us briefly highlight the low additional
computational load imposed by the refinement of the re-constructed
class subspace in the high-dimensional image space. Specifically,
note that the output of Steps~1 and 3 in Figure~\ref{f:summary} can
be pre-computed, as it is dependent only on the \emph{dimensions} of
the low and high scale data, not the data itself. Orthogonalization
in Step~2 is fast, as $D$ -- the number of columns in $\mathbf{B}_X$
-- is small. Although at first sight more complex, the
orthogonalization in Step~4 is also not demanding, as $\mathbf{B}_c$
is already orthonormal, so it is in fact only the $D$ columns of
$\mathbf{B}^*_X$ which need to be adjusted. Lastly, the Singular
Value Decomposition in Step~6 operates on a matrix which has a high
``landscape'' eccentricity so the first $D$ ``input'' directions can
be computed rapidly, while Step~7 consists only of a simple matrix
multiplication.

\section{Experimental Analysis}\label{s:exp}
The theoretical ideas put forward in the preceding sections were
evaluated empirically on two popular problems in computer vision:
matching sets of images of (i) face appearances and (ii) object
appearances. For this, two large data sets were used. These are:
\begin{itemize}
  \item The Cambridge Face Motion Database \cite{Aran2010b,Aran2012}
            \footnote{Also see \url{http://mi.eng.cam.ac.uk/~oa214/}.}, and\\

  \item The Amsterdam Library of Object Images \cite{GeusBurgSmeu2005}
            \footnote{Also see \url{www.science.uva.nl/~aloi/}.}.
\end{itemize}
Their contents are reviewed next.

\subsection{Data}
For a thorough description of the two data sets used, the reader should
consult previous publications in which they are described in detail,
respectively \cite{Aran2012} and \cite{GeusBurgSmeu2005}. Here they
are briefly summarized for the sake of clarity and completeness of the
present analysis.

\subsubsection{Cambridge Face Motion Database}
\begin{figure}
  \centering
  \footnotesize
  \begin{tabular}{c}
    \includegraphics[width=1.0\textwidth]{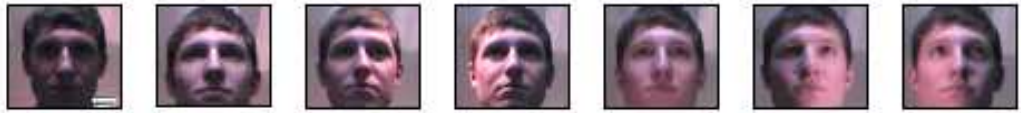}\\[-15pt]
    (a) Same pose and identity, different illuminations\\[5pt]
    \includegraphics[width=0.72\textwidth]{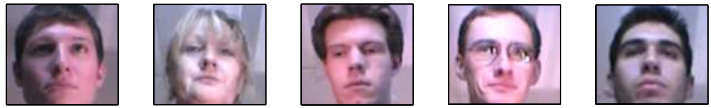}\\[-15pt]
    (b) Same illumination, different poses relative to the sources of illumination\\[10pt]
  \end{tabular}
  \caption{ (a) Illuminations 1--7 from the Cambridge face motion database.
            (b) Five different individuals in the illumination setting number 6. In spite of the
            same spatial arrangement of light sources, their effect on the appearance of
            faces changes significantly due to variations in people's heights and the
            \textit{ad lib} chosen position relative to the camera. }
            \label{f:illum}
  \vspace{30pt}
\end{figure}

The Cambridge Face data set is a database of face motion video
sequences acquired in the Department of Engineering, University of
Cambridge. It contains 100 individuals of varying age, ethnicity and
sex. Seven different illumination configurations were used for the
acquisition of data. These are illustrated in Figure~\ref{f:illum}.
For every person enrolled in the database 2 video sequences of the
person performing pseudo-random motion were collected in each
illumination. The individuals were instructed to approach the camera,
thus choosing their positioning \textit{ad lib}, and freely perform
head and/or body motion relative to the camera while real-time visual
feedback was provided on the screen placed above the camera. Most
sequences contain significant yaw and pitch variation, some
translatory motion and negligible roll. Mild facial expression changes
are present in some sequences (e.g.\ when the user was smiling or
talking to the person supervising the acquisition).

\subsubsection{Amsterdam Library of Object Images}
\begin{figure}[htb]
  \centering
  \subfigure[Object 0001 -- toy bear]{\includegraphics[width=1.00\textwidth]{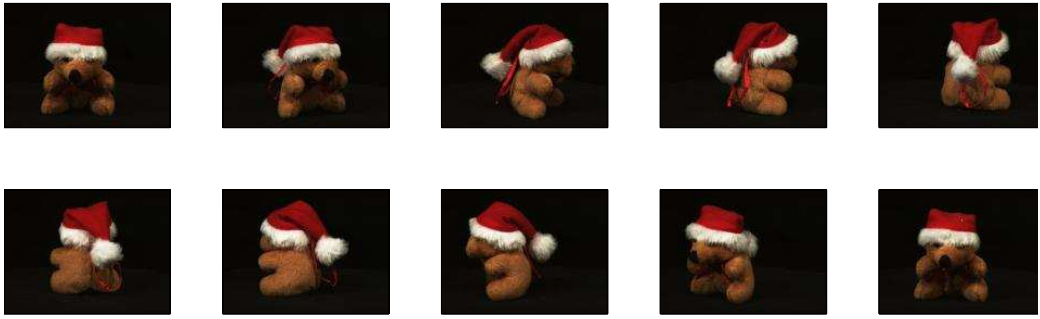}}
  \subfigure[Object 0002 -- keys on a chain]{\includegraphics[width=1.00\textwidth]{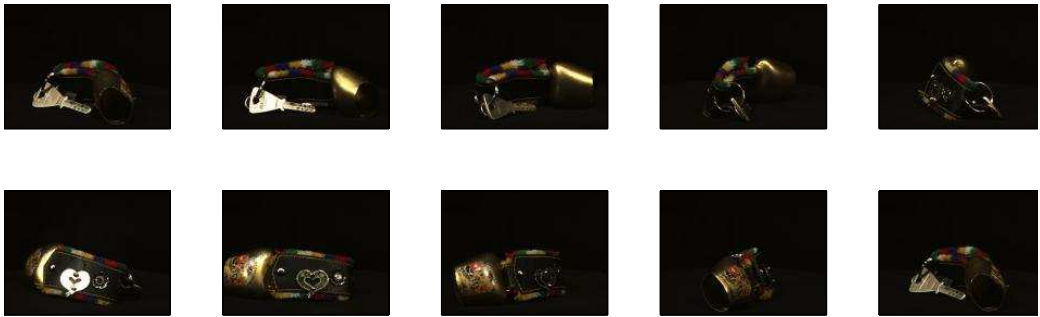}}
  \caption{ Examples of 10 roughly angularly equidistant views (out of 73 available) of two
            objects from the  ``Object View Collection'' subset of the Amsterdam Library of
            Object Images \cite{GeusBurgSmeu2005}.  }
  \label{f:aloi}
  \vspace{30pt}
\end{figure}

The Amsterdam Library of Object Images is a collection of images of 1000 small objects
\cite{GeusBurgSmeu2005}. Examples of two objects are shown in Figure~\ref{f:aloi}. The data set
comprises three main subsets: (i) ``Illumination Direction Collection'', (ii) ``Illumination
Colour Collection'' and (iii) ``Object View Collection''. In the ``Illumination Direction
Collection'' the camera viewpoint relative to each object was kept constant, while illumination
direction was varied. Similarly in the ``Illumination Colour Collection'', images corresponding
to different voltages of a variable voltage halogen illumination source were
acquired from a fixed viewpoint. Finally, ``Object View Collection'' contains view of objects
under a constant illumination but variable pose. These were acquired using 5$^{\circ}$ increments
of the object's rotation around an axis parallel to the image plane. This collection was used
in the evaluation reported here. Figure~\ref{f:aloi} shows a subset of 10 images (out of the
total number of $360/5+1 = 73$) which illustrate the nature of the data variability. Further
detail can be obtained by consulting the original publication \cite{GeusBurgSmeu2005} and from
the web site of the database: \url{www.science.uva.nl/~aloi/}.

\subsection{Evaluation Protocol}
\begin{figure}[htbp]
  \centering
  \subfigure[Face scales]{\includegraphics[width=0.80\textwidth]{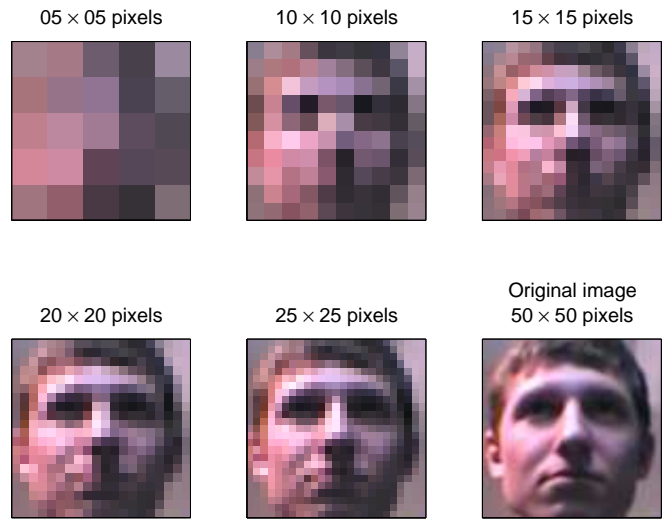}}
  \subfigure[Object scales]{\includegraphics[width=1.00\textwidth]{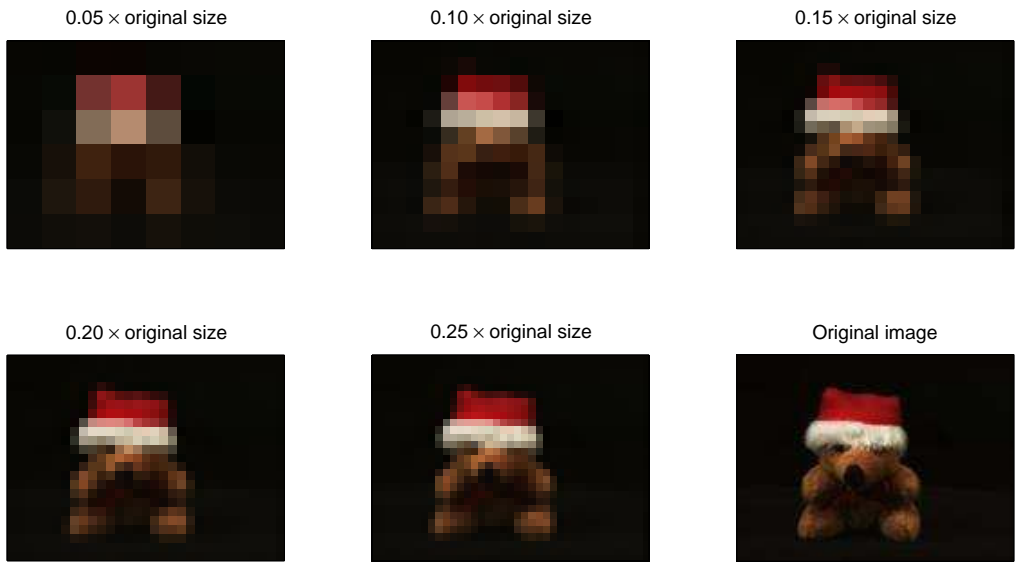}}
  \caption{Different scales used as low resolution matching input for (a) face and (b)
           object data. Square face images with the widths of 5 to 25 pixels at 5 pixel
           increments, were considered. Images from the Amsterdam Library of Object Images were
           sub-sampled to 5\% to 25\% of the original size, at 5\% increments. }
  \label{f:subs}
  \vspace{20pt}
\end{figure}

In the case of both data sets, evaluation was performed by matching high resolution with
low resolution class models. A single class was taken to correspond to a particular person
or an object when, respectively, face and object appearance was matched. High resolution
linear subspace models were computed using $50 \times 50$ pixel face data and $192 \times 144$
pixel object images, as described in Section~\ref{s:matching}. Low resolution subspaces were constructed
using downsampled data. Square face images were downsampled to five different scales: $5 \times 5$ pixels,
$10 \times 10$ pixels, $15 \times 15$ pixels, $20 \times 20$ pixels and $25 \times 25$ pixels,
as shown in Figure~\ref{f:subs}~(a). Data from the Amsterdam Library of Object Images
was downsampled also to five different scales corresponding to 5\%, 10\%, 15\%, 20\% and
25\% of its linear scale (e.g.\ height, while maintaining the original aspect ratio), as
shown in Figure~\ref{f:subs}~(b).

Training was performed by constructing class models with downsampled face images in a single
illumination setting in the case of face appearance matching and downsampled object images
using half of the available data in the case of object appearance matching.
Thus each class represented by a linear subspace corresponds to, respectively, a single
person and captures his/her appearance in the training illumination, and a single
object using a limited set of views.

In querying an algorithm using a novel subspace, the subspace was classified to the class of
the highest similarity. The similarity between two subspaces was expressed by a number in
the range $[0,1]$, equal to the correlation of the two highest correlated vectors confined
to them, as per Equation~\eqref{e:corr} in the previous section.

\subsection{Results}
First, the effects of the method proposed in Section~\ref{ss:proposed} on class separation were examined, and compared to that of the na\"{\i}ve method of Section~\ref{ss:naive}. This was quantified as follows. For a given pair of training and ``query'' illumination conditions, the similarity $\rho_{i,j}$ between all image sets $i$ acquired in the training illumination and all sets $j$ acquired in the query illumination was evaluated. Thus, the mean confidences $\overline{e}_w$ and $\overline{e}_b$ of, respectively, the correct and incorrect matching assignments are given by:
\begin{align}
  \overline{e}_w &= 1 - \frac{1}{M}~\sum_{i=1}^{M} \rho_{i,i} \\
  \notag\\
  \overline{e}_b &= 1 - \frac{1}{M \times (M-1)}~\sum_{i=1}^{M} \sum_{\scriptsize
    j=1,~j \neq i}^{M}  \rho_{i,j},
\end{align}
where $M$ is the number of distinct classes. The corresponding separation is then proportional to $\overline{e}_b$ and inversely proportional to $\overline{e}_w$:
\begin{align}
  \mu = \overline{e}_b~{\overline{e}_w}^{-1}.
\end{align}

\begin{figure}[htbp]
  \centering
  \subfigure[$(5 \times  5) \longleftrightarrow (50\times 50)$]{\includegraphics[width=0.48\textwidth]{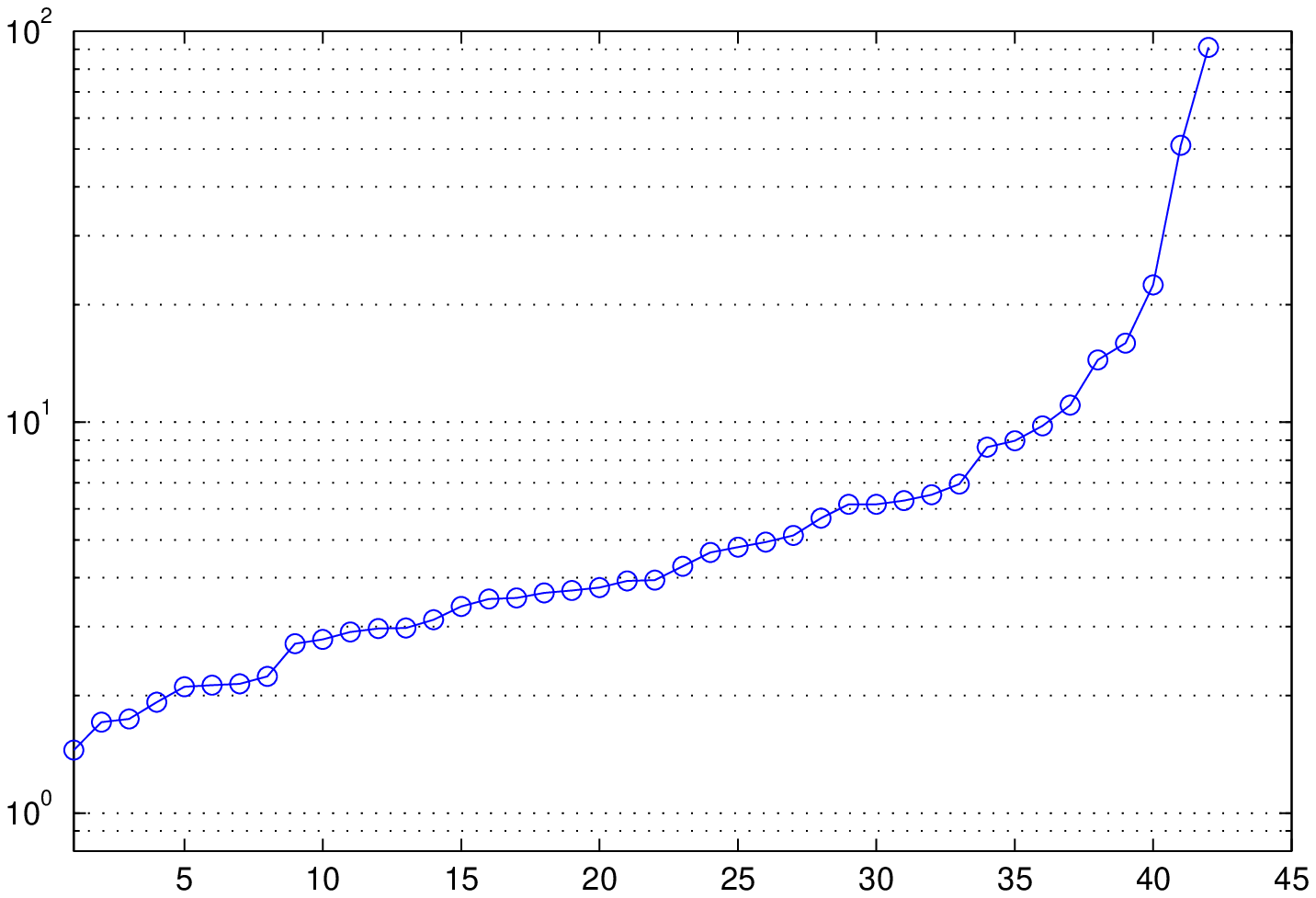}}
  \subfigure[$(10\times 10) \longleftrightarrow (50\times 50)$]{\includegraphics[width=0.48\textwidth]{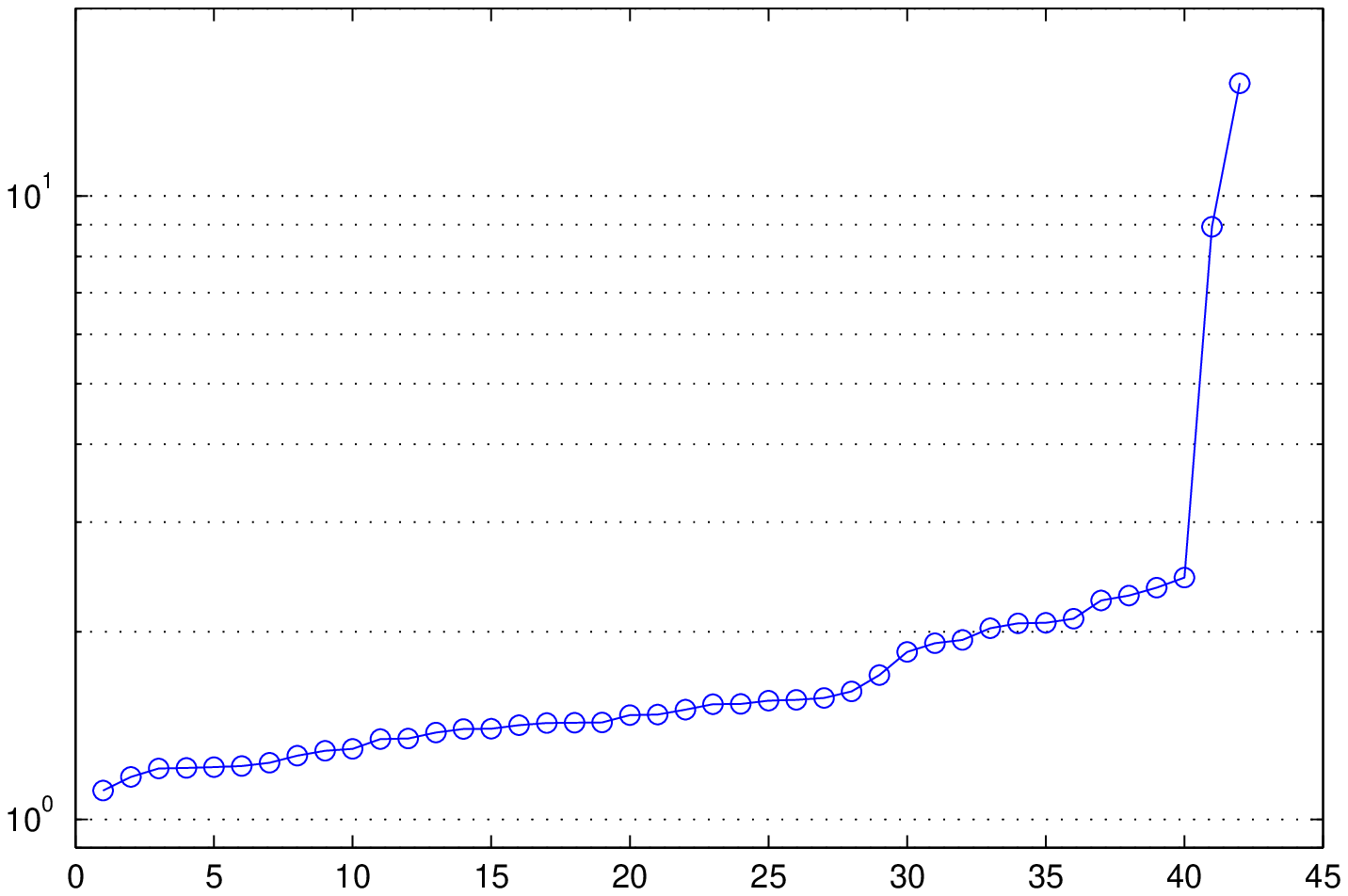}}
  \vspace{5pt}
  \subfigure[$(15\times 15) \longleftrightarrow (50\times 50)$]{\includegraphics[width=0.48\textwidth]{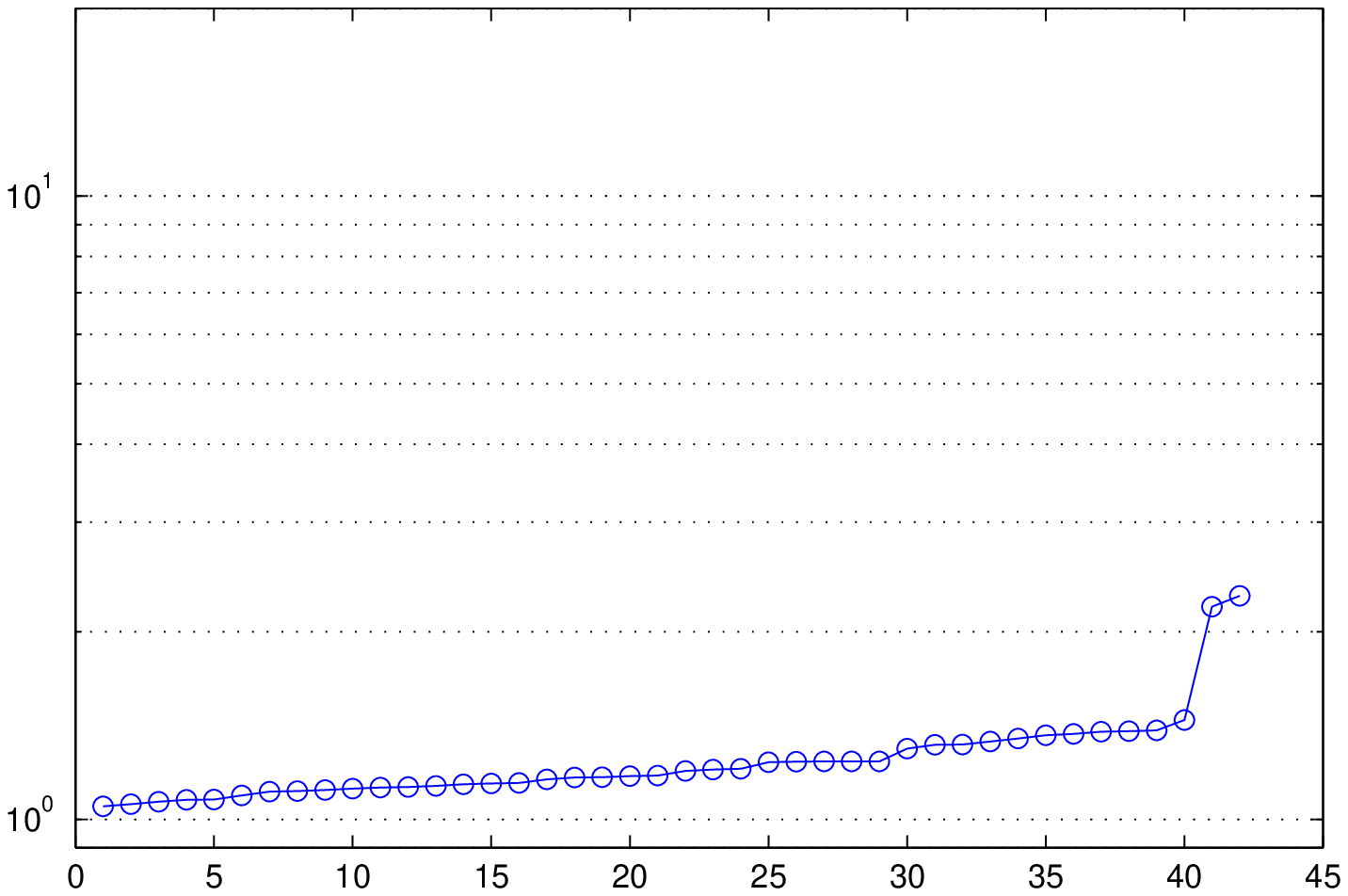}}
  \subfigure[$(20\times 20) \longleftrightarrow (50\times 50)$]{\includegraphics[width=0.48\textwidth]{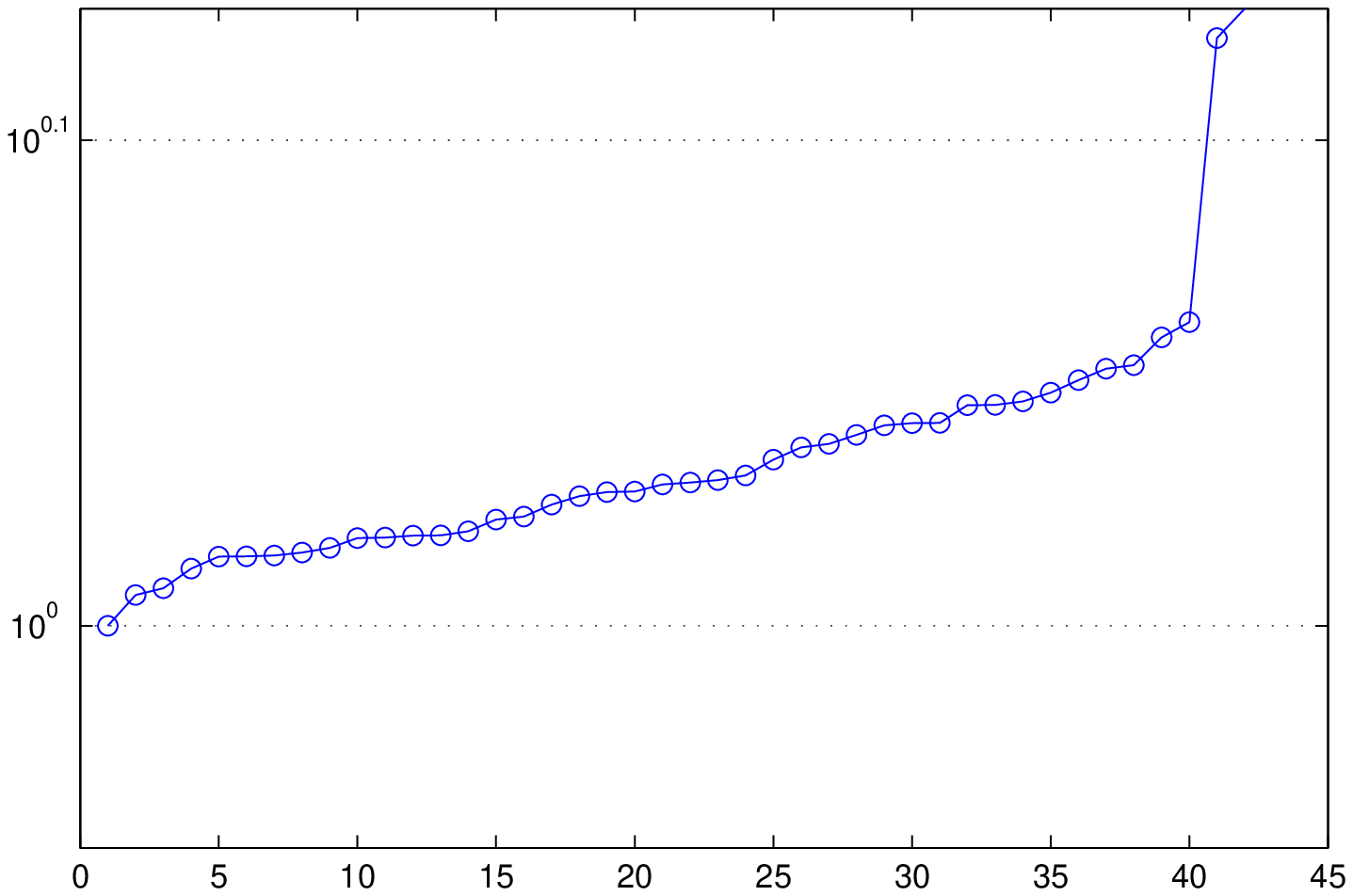}}
  \vspace{10pt}
  \caption{The increase in class separation $\mu$ (the ordinate; note that the scale is logarithmic) over
           different training-query illumination conditions (abscissa), achieved by the proposed method
           in comparison to the na\"{\i}ve subspace re-projection approach. Note that for clarity the
           training-query illumination pairs were ordered in increasing order of improvement for each plot; thus,
           the indices of different abscissae do not necessarily correspond. }
  \label{f:sep}
  \vspace{20pt}
\end{figure}

The separation was evaluated separately for all training-query illumination pairs in the Cambridge Face Database using the na\"{\i}ve method and compared with that of the proposed solution across different matching scales using the bicubic projection model. A plot of the results is shown in Figure~\ref{f:sep} in which for clarity the training-query illumination pairs were ordered in increasing order of improvement for each plot (thus the indices of different abscissae do not necessarily correspond).

Firstly, note that improvement was observed for all illumination combinations at all scales. Unsurprisingly, the most significant increase in class separation ($\approx 8.5$-fold mean increase) was achieved for the most drastic difference in training and query sets, when subspaces embedded in a 25-dimensional image space -- representing the appearance variation of images as small as $5 \times 5$ pixels, see Figure~\ref{f:subs}~(a) -- was matched against a subspace embedded in the image space of a 100 times greater dimensionality.

\begin{figure}[htbp]
  \centering
  \footnotesize
  \begin{tabular}{cc}
    \multicolumn{2}{c}{\includegraphics[width=1.00\textwidth]{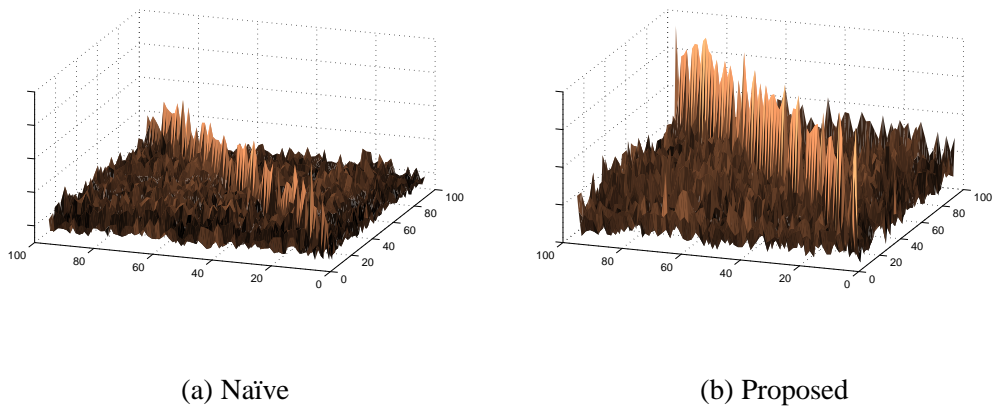}}\\
    \hspace{80pt}(a) Na\"{\i}ve & \hspace{80pt}(b) Proposed\\
  \end{tabular}
  \caption{Typical similarity matrices resulting from the na\"{\i}ve (left) and the proposed (right) matching approaches.
            Our method produces a dramatic improvement in class separation as witnessed by the increased dominance
            of the diagonal elements in the aforementioned matrix. }
  \label{f:distmat}
  \vspace{20pt}
\end{figure}

It is interesting to note that even at the more favourable scales of the low resolution input, although the mean improvement was less noticeable than at extreme scale discrepancies, the accuracy of matching in certain combinations of illumination settings still greatly benefited from the proposed method. For example, for low resolution subspaces representing appearance in $10 \times 10$ pixel images, the mean separation increase of 75.6\% was measured; yet, for illuminations ``1'' and ``2'' -- corresponding to the index 42 on the abscissa in Figure~\ref{f:sep}~(b) -- the improvement was 473.0\%. The change effected on the inter-class and intra-class distances is illustrated in Figure~\ref{f:distmat}, which shows a typical similarity matrix produced by the na\"{\i}ve and the proposed matching methods.

\begin{figure}[htbp]
  \centering
  \subfigure[Faces -- bilinear projection model]{\includegraphics[width=0.9\textwidth]{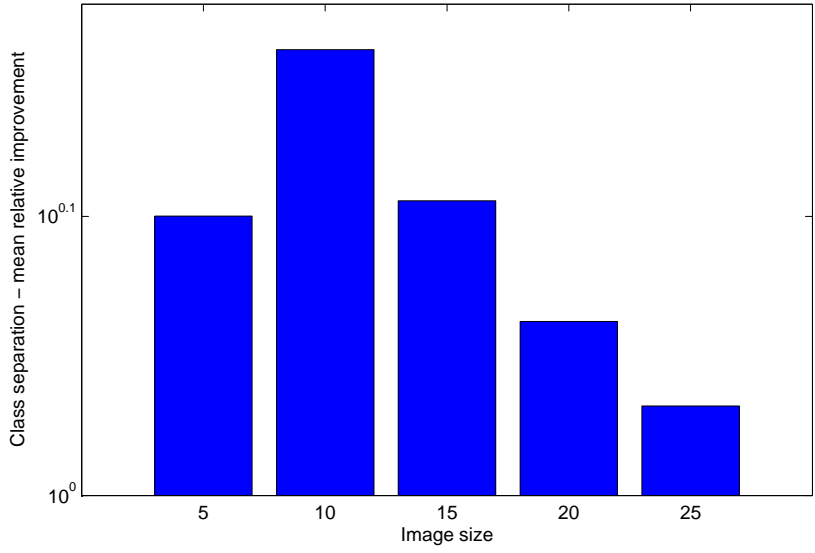}}
  \subfigure[Faces -- bicubic projection model]{\includegraphics[width=0.9\textwidth]{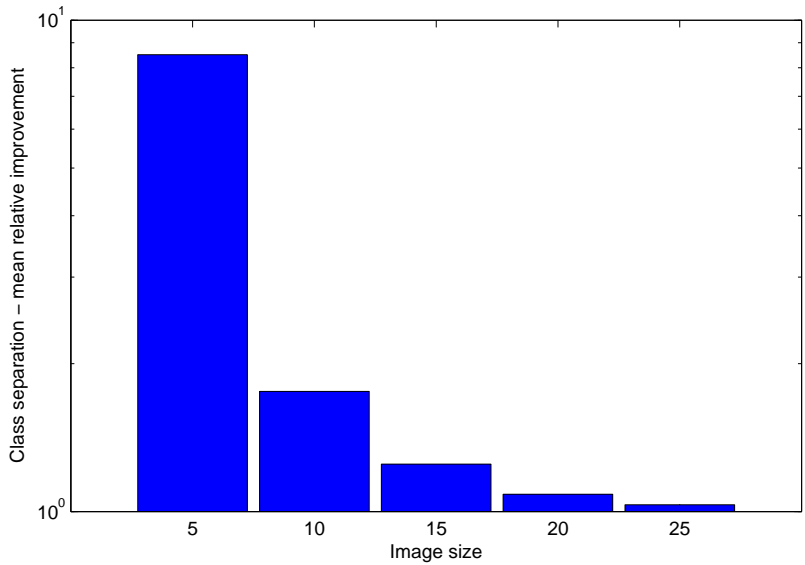}}
  \vspace{10pt}
  \caption{ Mean class separation increase achieved, as a function of size of the low-scale
            images. Shown is the ratio of class separation when subspaces are matched using the proposed method and the
            na\"{\i}ve re-projection method described in Section~\ref{ss:naive}. The rate of improvement decay is
            incrementally exponential, reaching 1 (no improvement) when $d_l = d_h$. }
  \label{f:imp}
  \vspace{20pt}
\end{figure}

\begin{figure}[htbp]
  \centering
  \subfigure[Objects -- bilinear projection model]{\includegraphics[width=0.9\textwidth]{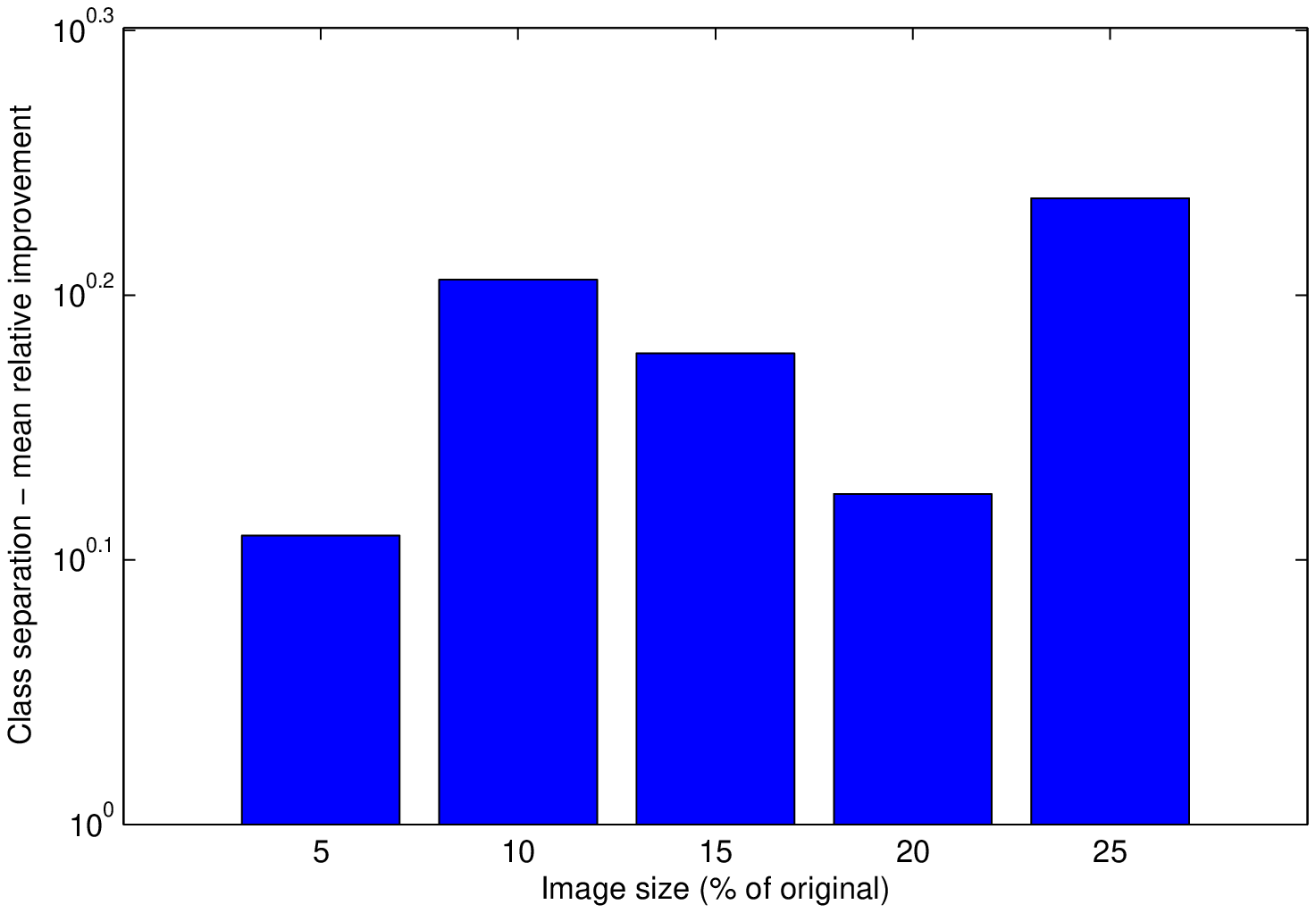}}
  \subfigure[Objects -- bicubic projection model]{\includegraphics[width=0.9\textwidth]{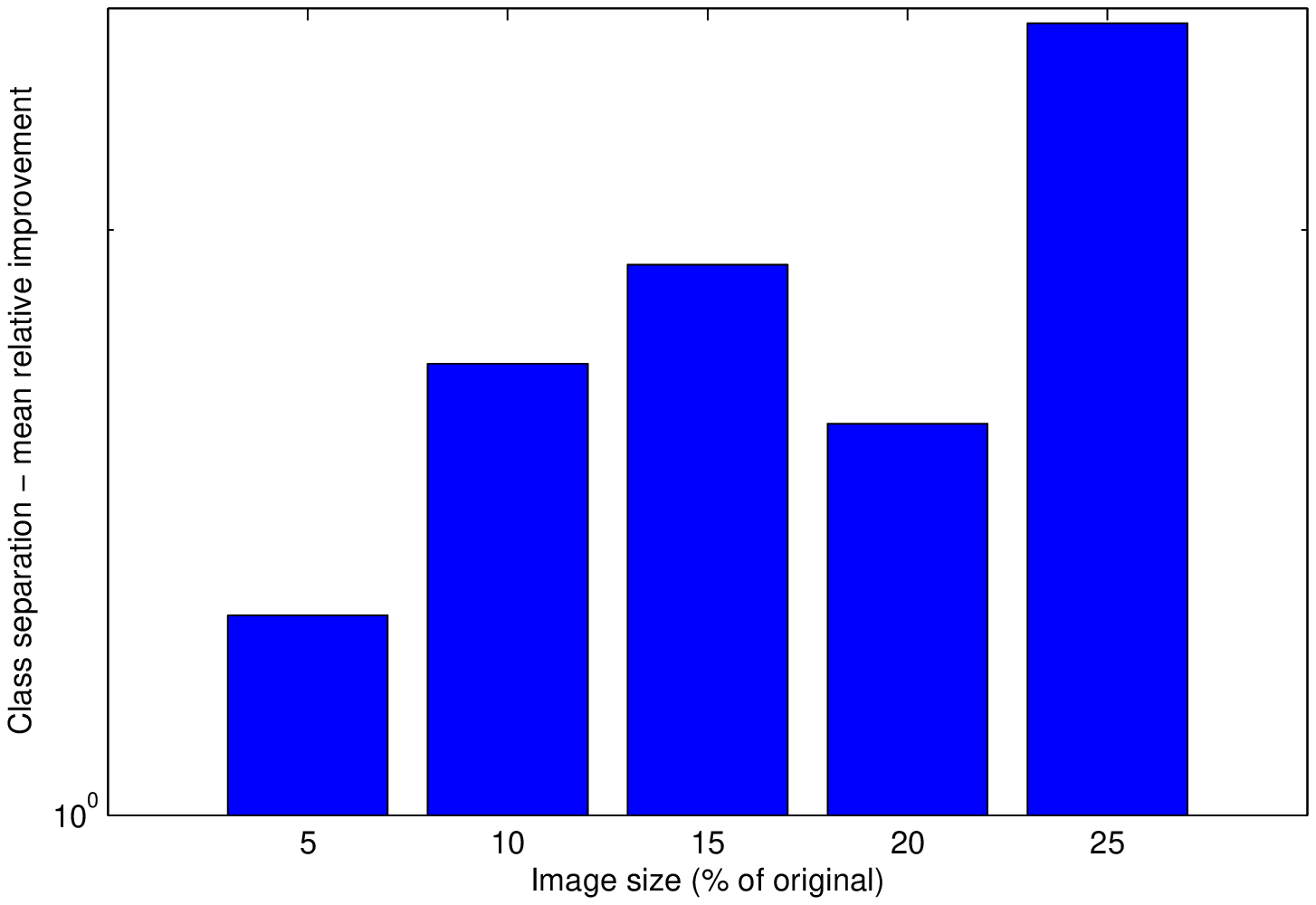}}
  \vspace{10pt}
  \caption{ Mean class separation increase achieved, as a function of size of the low-scale
            images. Shown is the ratio of class separation when subspaces are matched using the proposed method and the
            na\"{\i}ve re-projection method. Unlike in the plots obtained from face matching experiments in
            Figure~\ref{f:imp}, the nature of variation across different scales here appears less regular. It is likely
            that the reason lies in the large (and variable in shape) area of the background present in
            the object images. }
  \label{f:impALOI}
  \vspace{20pt}
\end{figure}

The mean separation increase across different scales for face and object
data is shown in, respectively, Figures~\ref{f:imp} and~\ref{f:impALOI}.
These also illustrate the impact that the projection model used has on the
quality of matching results,
Figures~\ref{f:imp}~(a) and~\ref{f:impALOI}~(a) corresponding to the
bilinear projection model, and Figures~\ref{f:imp}~(b) and~\ref{f:impALOI}~(b)
to the bicubic. As could be expected from theory, the latter
was found to be consistently superior across all scales and for both data sets.
In the case of face appearance, the greatest improvement over the
na\"{\i}ve re-projection method was observed for the smallest scale of
low resolution data -- $8.5$-fold separation increase was achieved
for $5 \times 5$ pixel images, $1.75$-fold for $10 \times 10$,
$1.25$-fold for $15 \times 15$, $1.08$-fold for $20 \times 20$ and $1.03$-fold
for $25 \times 25$. It is interesting to note that the relative performance
across different scales of low resolution object data did not follow the
same functional form as in the case of face data. A possible reason for this
seemingly odd result lies in the
presence of confounding background regions (unlike in the face data set, which was
automatically cropped to include foreground information only). Not only does
the background typically occupy a significant area of object images, it is also
of variable shape across different views of the same object, as well as
across different objects. It is likely that the interaction of this confounding
factor with the downsampling scale is the cause of the less predictable
nature of the plots in Figure~\ref{f:impALOI}.

\begin{figure}[htbp]
  \centering
  \subfigure[$(5\times 5)   \longleftrightarrow (50\times 50)$]{\includegraphics[width=0.44\textwidth]{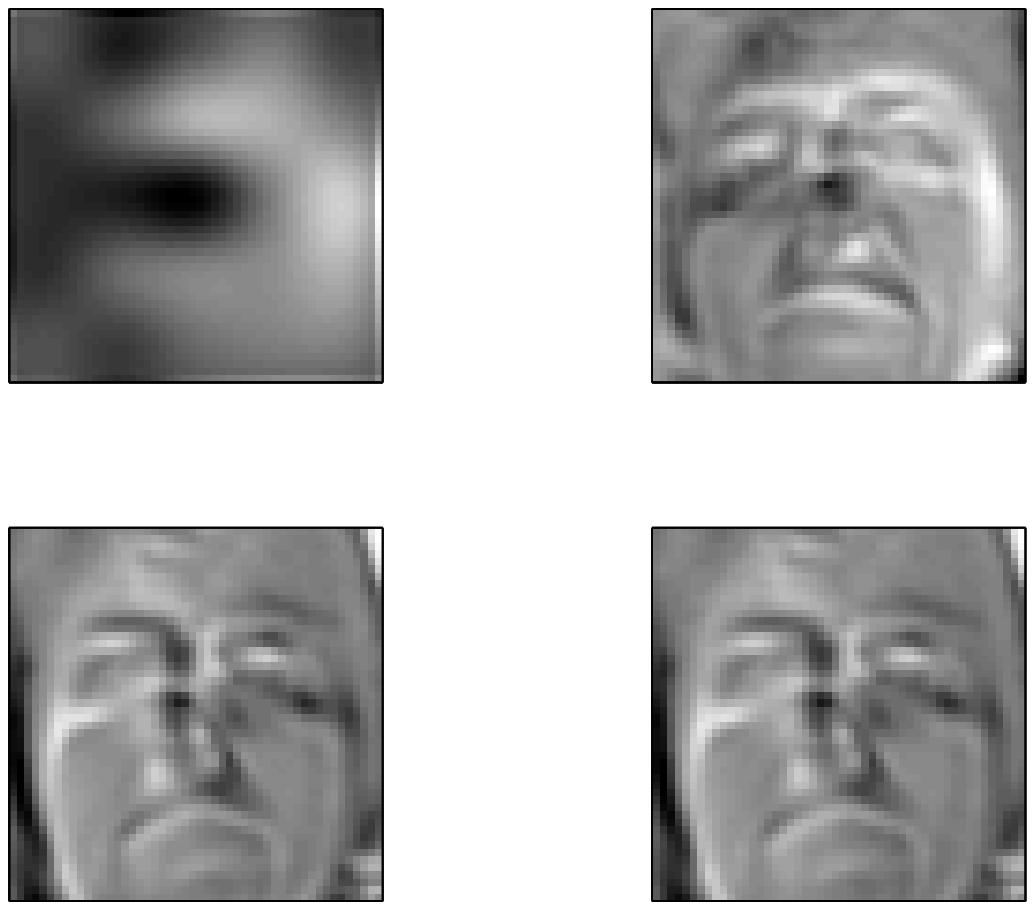}}
  \subfigure[$(10\times 10) \longleftrightarrow (50\times 50)$]{\includegraphics[width=0.44\textwidth]{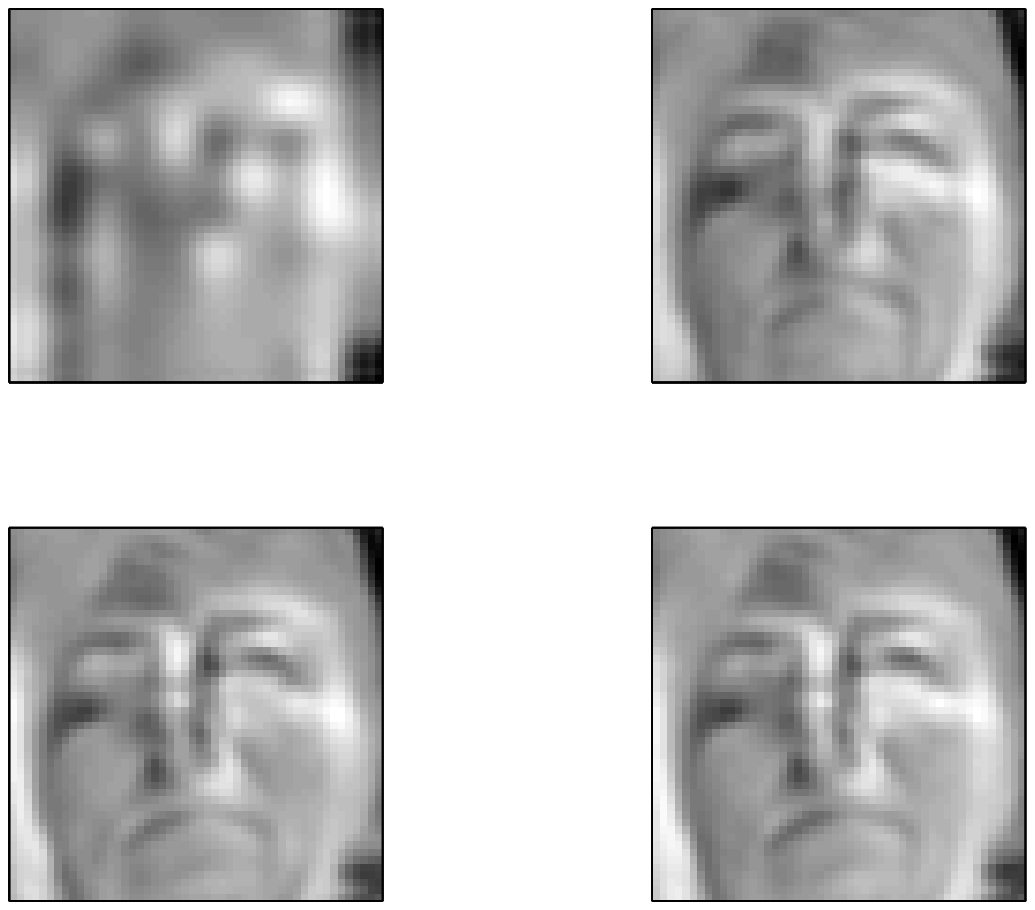}}
  \subfigure[$(15\times 15) \longleftrightarrow (50\times 50)$]{\includegraphics[width=0.44\textwidth]{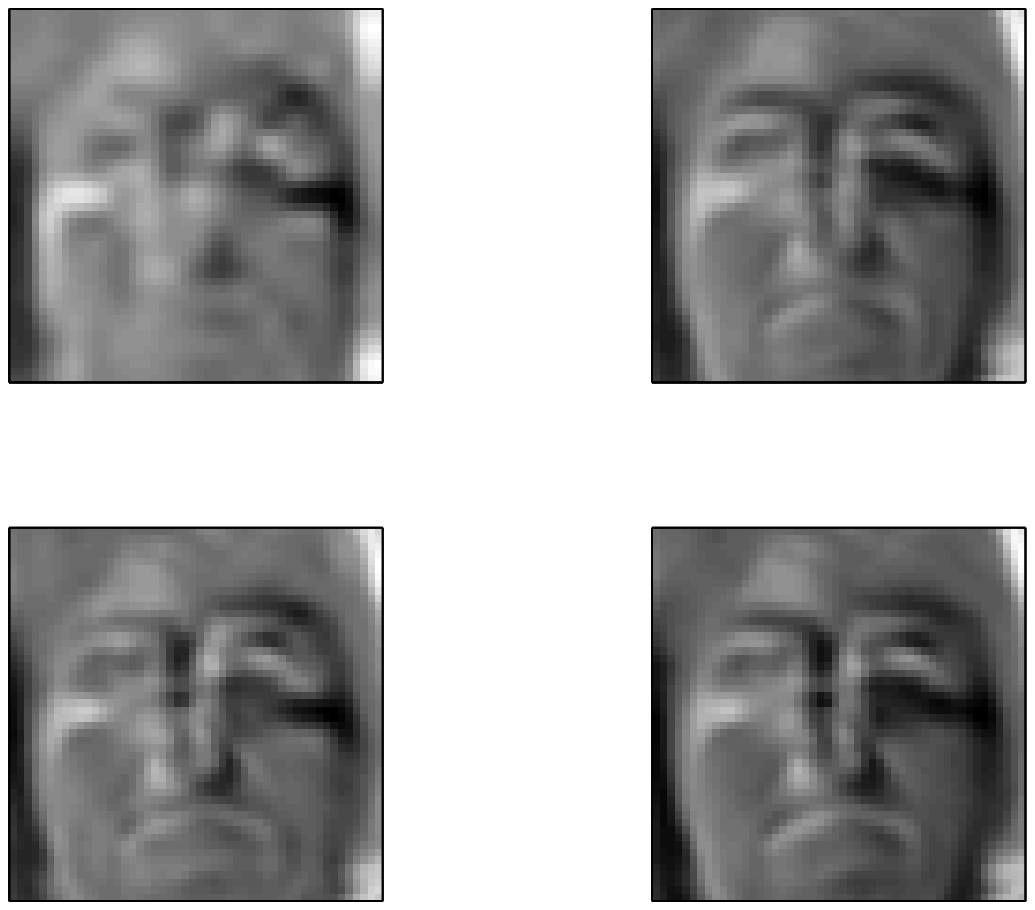}}
  \subfigure[$(20\times 20) \longleftrightarrow (50\times 50)$]{\includegraphics[width=0.44\textwidth]{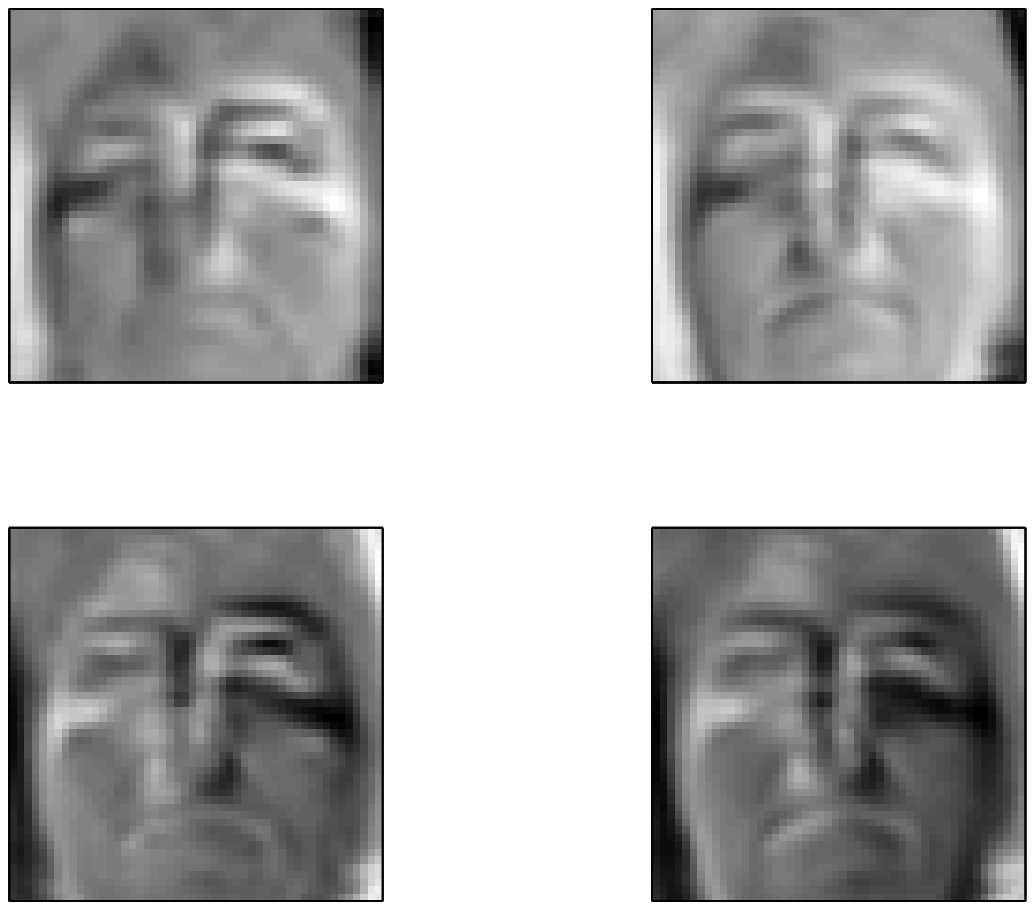}}
  \subfigure[$(25\times 25) \longleftrightarrow (50\times 50)$]{\includegraphics[width=0.44\textwidth]{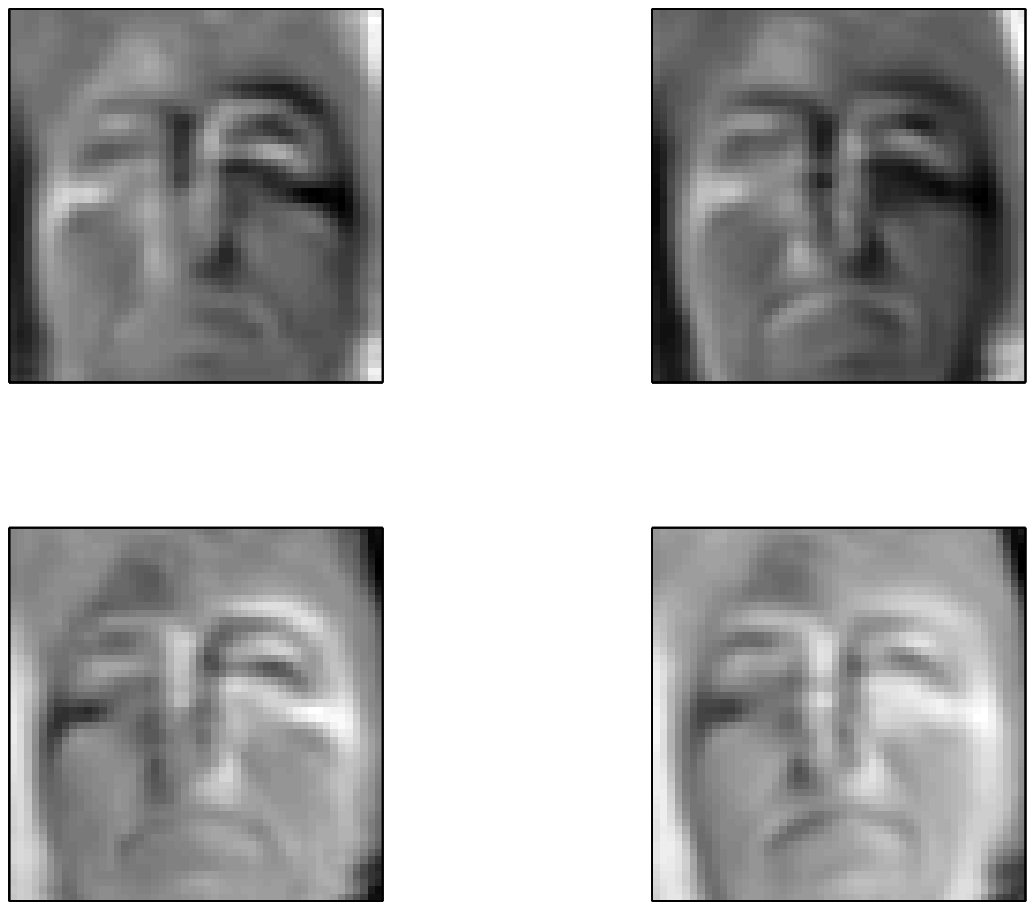}}
  \caption{ Bicubic projection model -- the inferred most similar modes of variation contained within two subspaces representing
            face appearance variation of the same person in different illumination conditions and at different training scales.
            In each subfigure, which corresponds to a different training-query scale discrepancy, the top pair of images
            represents appearance extracted by the na\"{\i}ve algorithm of Section~\ref{ss:naive} (as the left-singular and right-singular
            vectors of ${\mathbf{B}_Y}^T~\mathbf{B}^*_X$); the bottom pair is extracted by the proposed method (as the
            left-singular and right-singular vectors of ${\mathbf{B}_Y}^T~\mathbf{B}_{Xc}$). }
  \label{f:modes1}
  \vspace{20pt}
\end{figure}

\begin{figure}[htbp]
  \centering
  \subfigure[$(5\times 5)   \longleftrightarrow (50\times 50)$]{\includegraphics[width=0.44\textwidth]{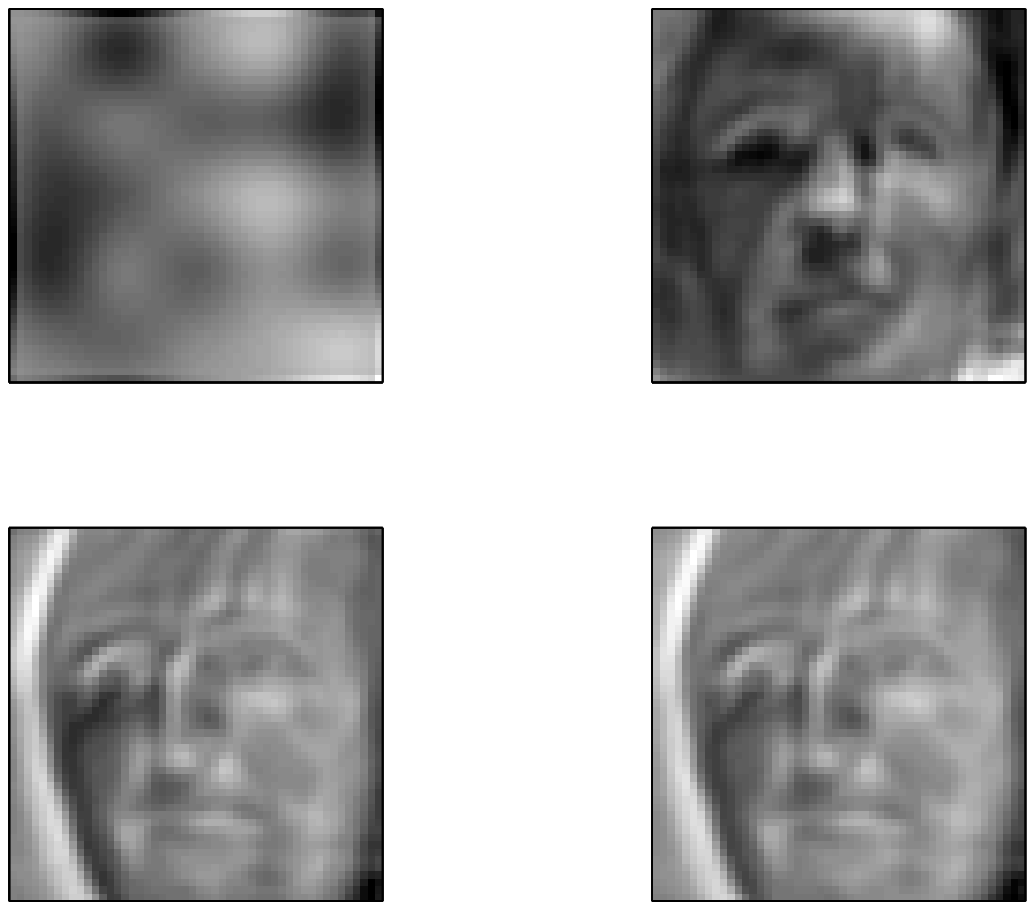}}
  \subfigure[$(10\times 10) \longleftrightarrow (50\times 50)$]{\includegraphics[width=0.44\textwidth]{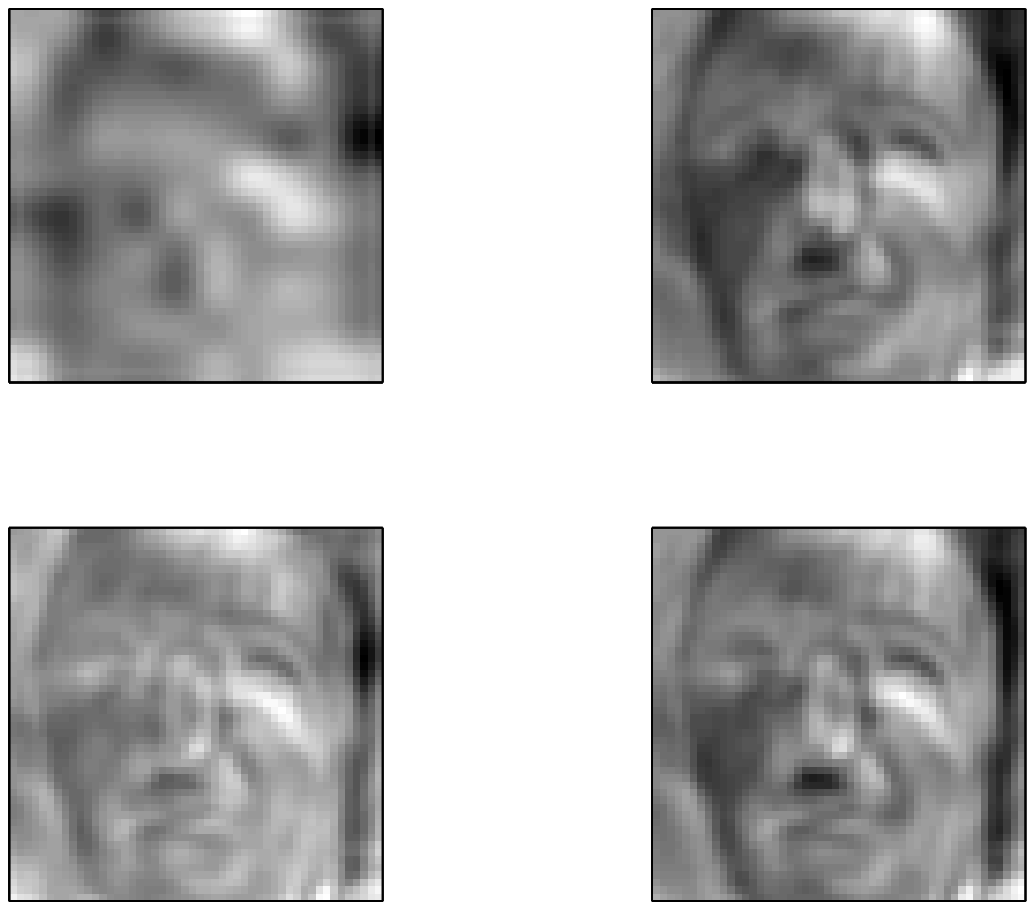}}
  \subfigure[$(15\times 15) \longleftrightarrow (50\times 50)$]{\includegraphics[width=0.44\textwidth]{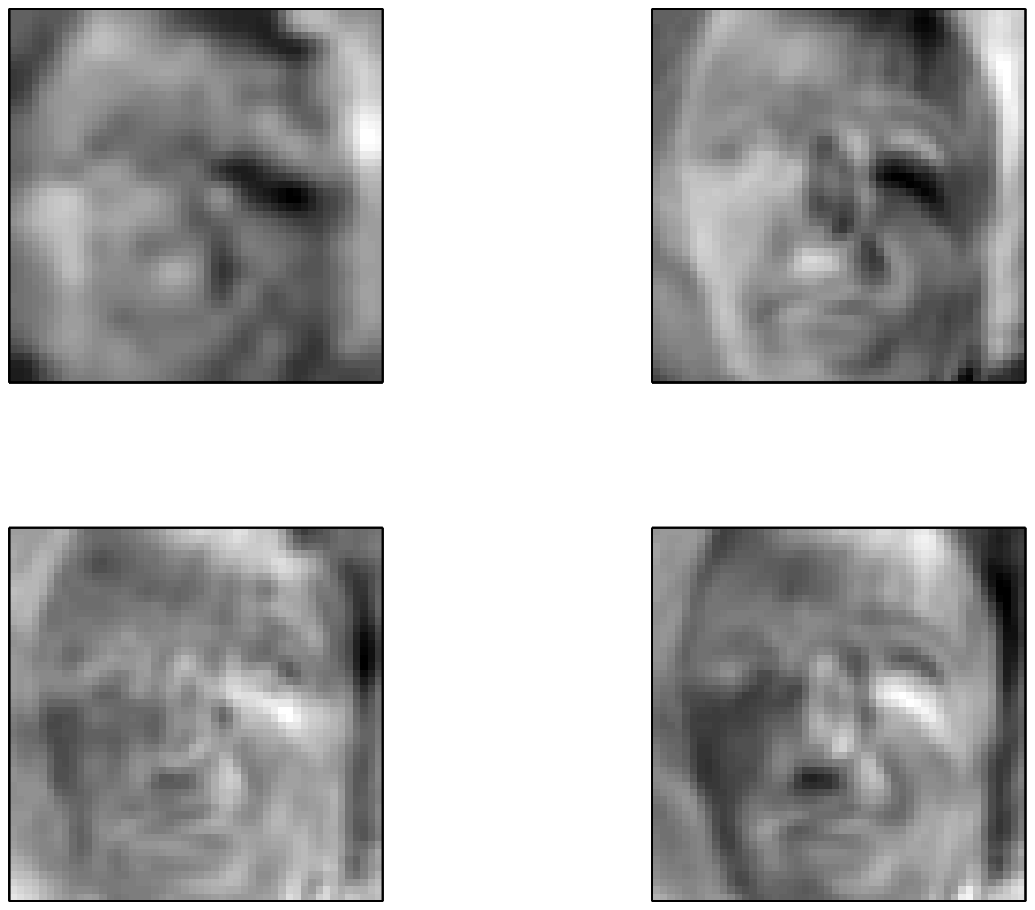}}
  \subfigure[$(20\times 20) \longleftrightarrow (50\times 50)$]{\includegraphics[width=0.44\textwidth]{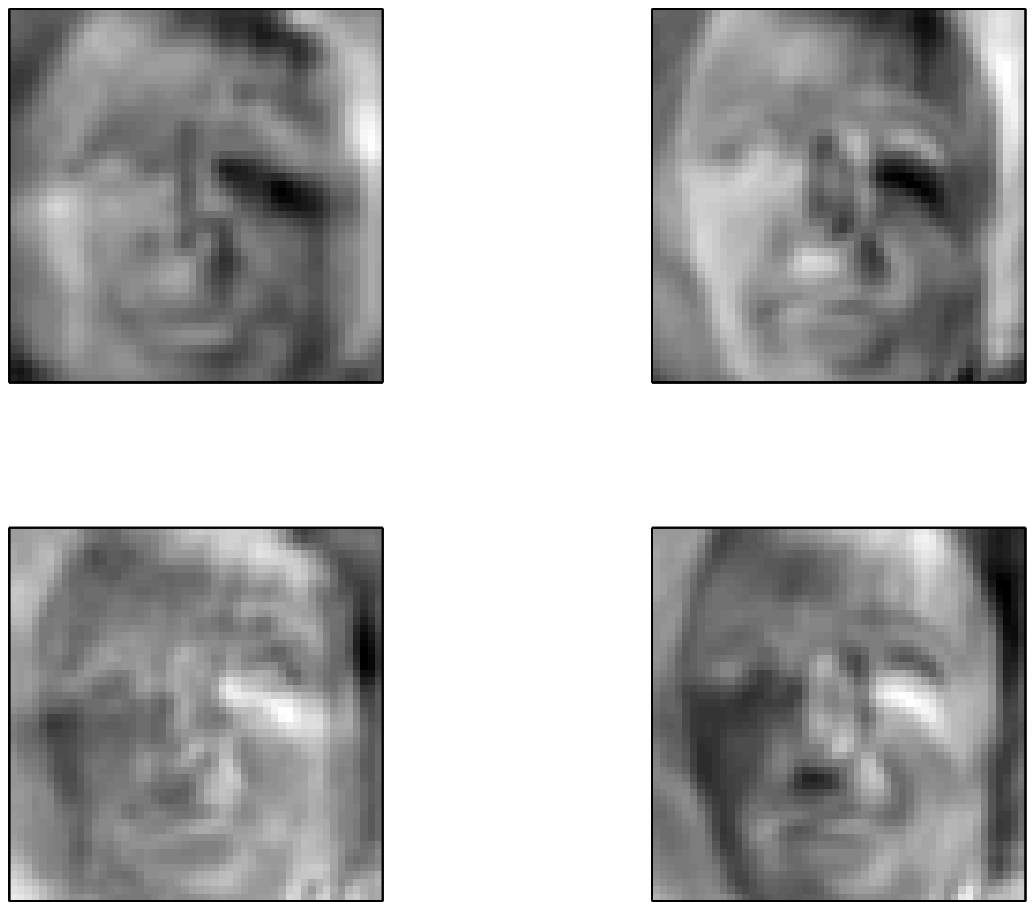}}
  \subfigure[$(25\times 25) \longleftrightarrow (50\times 50)$]{\includegraphics[width=0.44\textwidth]{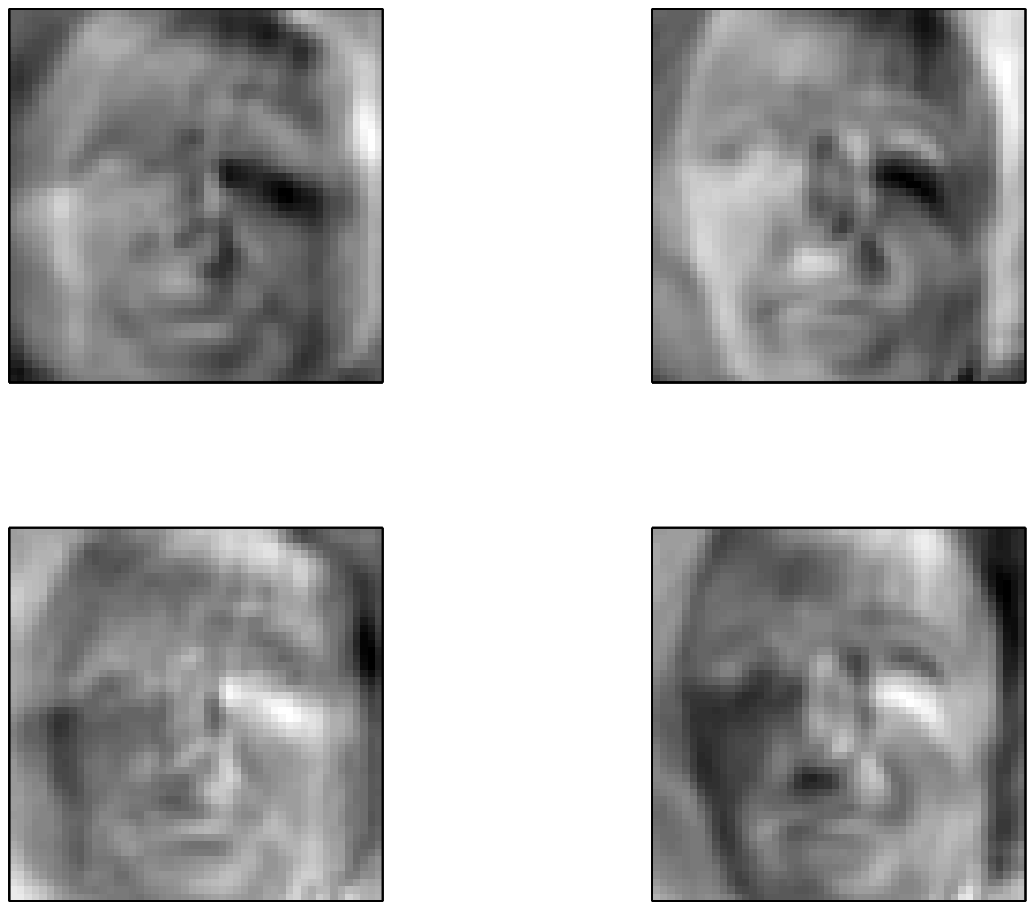}}
  \caption{ Bilinear projection model -- the inferred most similar modes of variation contained within two subspaces representing face appearance
            variation of the same person in different illumination conditions and at different training scales.
            In each subfigure, which corresponds to a different training-query scale discrepancy, the top pair of images
            represents appearance extracted by the na\"{\i}ve algorithm of Section~\ref{ss:naive} (as the left-singular and right-singular
            vectors of ${\mathbf{B}_Y}^T~\mathbf{B}^*_X$); the bottom pair is extracted by the proposed method (as the
            left-singular and right-singular vectors of ${\mathbf{B}_Y}^T~\mathbf{B}_{Xc}$). }
  \label{f:modes2}
  \vspace{20pt}
\end{figure}

The inferred most similar modes of variation contained within two subspaces representing face appearance variation of the same person in different illumination conditions and at different
training scales for the bilinear and bicubic models respectively are shown in Figures~\ref{f:modes1} and~\ref{f:modes2}. In both cases, as the scale of low-resolution images is reduced, the na\"{\i}ve algorithm of Section~\ref{ss:naive} finds progressively worse matching modes with significant visual degradation in the mode corresponding to the low-resolution subspace. In contrast, the proposed algorithm correctly reconstructs meaningful high-resolution appearance even in the case of extremely low resolution images ($5 \times 5$ pixels).

\begin{figure}[htb]
  \centering
  \vspace{20pt}
  \includegraphics[width=0.90\textwidth]{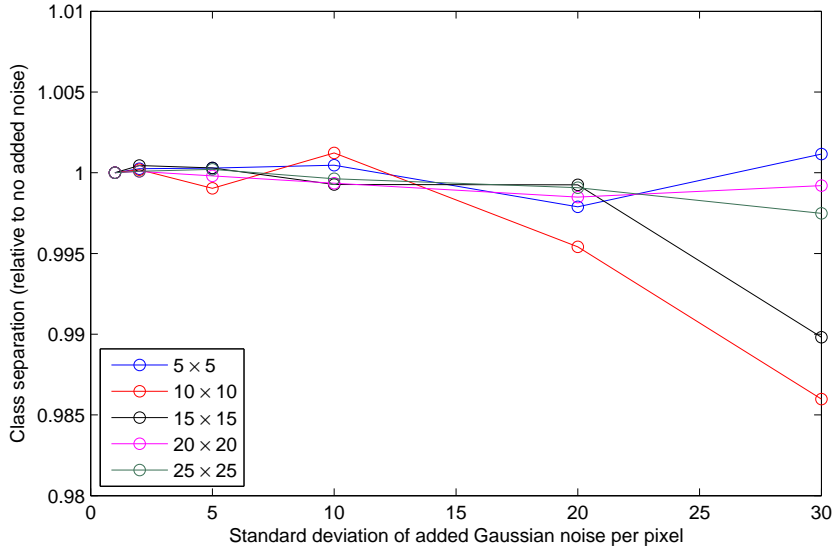}
  \caption{ The effects of additive zero-mean Gaussian noise, isotropic in the image space, applied to low resolution images before the construction of the corresponding subspaces. Shown is the change in the observed class separation which is for the sake of visualization clarity measured relative to the separation achieved using the original, un-corrupted images; see Figure~\ref{f:imp}(a). The results are for the bilinear projection model. Notice the remarkable robustness of the proposed model: even for noise with the pixel-wise standard deviation of 30 greyscale levels (approximately 12\% of the entire greyscale intensity range), which corresponds to the average signal-to-noise ratio of 1.7, class separation is decreased by less than 1.5\%.
  Note that this means that even when the proposed method performs matching in the presence of extreme noise, its performance exceeds that of the na\"{\i}ve approach applied to un-corrupted data. }
  \label{f:noise}
  \vspace{20pt}
\end{figure}

Lastly, we examined the behaviour of the proposed method in the presence of data corruption by noise. Specifically, we repeated the previously described experiments for the bilinear projection model with the difference that following the downsampling of high resolution images we added pixel-wise Gaussian noise to the resulting low resolution images before creating the corresponding low-dimensional subspaces. Since pixel-wise characteristics of noise were the same across all pixels in a specific experiment, this noise is isotropic in the low-dimensional image space. The sensitivity of the proposed method was evaluated by varying the magnitude of noise added in this manner. In particular, we started by adding noise with pixel-wise standard deviation of 1 (i.e.\ image space root mean square equal to $\sqrt{D_l}$) , or approximately 0.4\% of the entire greyscale spanning the range from 0 to 255, and progressively increased up to 30 (i.e.\ image space root mean square equal to $30\sqrt{D_l}$), or approximately 12\% of the possible pixel value range which corresponds to the average signal-to-noise ratio of only $1.7$. The results are summarized in the plot in Figure~\ref{f:noise} which shows the change in class separation for different levels of additive noise. Note that for the sake of easier visualization in a single plot, the change is shown relative to the separation attained using un-corrupted images, discussed previously and plotted in Figure~\ref{f:imp}(a). It is remarkable to observe that even in the most challenging experiment, when the magnitude of added noise is extreme, the performance of the proposed method is hardly affected at all. In all cases, including that when matching is performed using low-dimensional subspaces with the greatest downsampling factor, the average class separation is not decreased more than 1.5\%. For pixel-wise noise magnitudes of up to 20 greyscale levels, the deterioration is consistently lower than 0.5\%, and even for the pixel-wise noise magnitude of 30 greyscale levels the separation decrease of more than 1\% is observed in only two instances (for low-dimensional spaces corresponding to images downsampled to $10 \times 10$ and $15 \times 15$ pixels). Note that this means that even when the proposed method performs matching in the presence of extreme noise, its performance exceeds that of the na\"{\i}ve approach applied to un-corrupted data.

\section{Conclusion}\label{s:conclusion}
In this paper a method for matching linear subspaces which represent appearance variations in images of different scales was described. The approach consists of an initial re-projection of
the subspace in the low-dimensional image space to the high-dimensional one, and subsequent refinement of the re-projection through a constrained rotation. Using facial and object appearance images and the corresponding two large data sets, it was shown that the proposed algorithm successfully reconstructs the personal subspace in the high-dimensional image space even for low-dimensional input corresponding to images as small as $5 \times 5$ pixels, improving average class separation by an order of magnitude. Our immediate future work will be in the direction of integrating the proposed method with the discriminative framework recently described in \cite{Aran2013e}.

\section*{Acknowledgements} The author would like to thank Trinity
College Cambridge for their kind support and the volunteers from the University of Cambridge Department of Engineering whose face data was included in the database used in developing the algorithm described in this paper.

\bibliographystyle{unsrt}
\bibliography{./my_bibliography}
\end{document}